\documentclass[journal]{IEEEtran}

\usepackage{cite}

\usepackage[cmex10]{amsmath}
\usepackage{algorithmic}

\usepackage{array}
\usepackage{amsfonts,amssymb}
\usepackage{graphicx,float}
\usepackage{url,tabularx}
\usepackage{subfigure}
\usepackage{bm,amssymb}
\usepackage{accents}
\usepackage{amsthm}
\usepackage{comment}
\usepackage{color}

\newcommand{\mb}[1]{\mathbf{#1}}
\newcommand{\mc}[1]{\mathcal{#1}}
\newcommand{\tb}[1]{\textbf{#1}}
\newcommand{\mbb}[1]{\mathbb{#1}}

\definecolor{lightgray}{gray}{0.95} 

\newcommand{\Om}{\bm{\Omega}}
\newcommand{\Sig}{\bm{\Sigma}}
\newcommand{\Mat}{\mb{X}}
\newcommand{\Samp}{\mb{S}}
\newcommand{\X}{\mb{X}}

\newcommand{\hOmLon}{\widehat{\Om}_1}
\newcommand{\Uij}{\mb{U}_{ij}}
\newcommand{\Vij}{\mb{U}_{ji}}
\newcommand{\Id}{\mb{I}}
\newcommand{\Y}{\mb{Y}}

\newcommand{\Del}{\bm{\Delta}}
\newcommand{\hSigLon}{\widehat{\Sig}_1}
\newcommand{\Xo}{\X_0} 
\newcommand{\Yo}{\Y_0} 

\newcommand{\Z}{\mb{Z}}
\newcommand{\Zij}{\Z_{0,ij}}
\newcommand{\Zii}{\Z_{0,ii}}

\newcommand{\Xp}{\X^\prime}

\newcommand{\Xd}{\X^\bullet}

\newcommand{\hOm}{\widehat{\Om}}
\newcommand{\hOmav}{\hOm_{\textup{av}}}
\newcommand{\hOmavL}[1]{\hOm_{\textup{av},l_{#1}}}

\newcommand{\Xk}{\X^k}

\newcommand{\Xkp}{\X^{k+1}}
\newcommand{\Yk}{\Y^k}
\newcommand{\Ykp}{\Y^{k+1}}

\newcommand{\Omij}{\omega_{ij}}

\newcommand{\Sampij}{s_{ij}}
\newcommand{\Sampii}{s_{ii}}
\newcommand{\Mij}{x_{ij}}
\newcommand{\lam}{\lambda}
\newcommand{\delij}{\delta_{ij}}
\newcommand{\delii}{\delta_{ii}}
\newcommand{\Yij}{y_{ij}}
\newcommand{\Yii}{y_{ii}}
\newcommand{\Yjj}{y_{jj}}
\newcommand{\shurij}{\Delta_{ij}}
\newcommand{\shuroij}{\Delta_{0,ij}}
\newcommand{\del}{\delta}

\newcommand{\Xij}{x_{ij}}
\newcommand{\Xii}{x_{ii}}

\newcommand{\hOmLonij}{\widehat{\omega}_{1,ij}}
\newcommand{\hSigLonij}{\widehat{\sigma}_{1,ij}}
\newcommand{\Xoij}{x_{0,ij}} 
\newcommand{\Xoii}{x_{0,ii}} 

\newcommand{\Xsii}{m_{ii}}
\newcommand{\Xsij}{m_{ij}}
\newcommand{\Yoii}{y_{0,ii}} 
\newcommand{\Yojj}{y_{0,jj}} 
\newcommand{\Yoij}{y_{0,ij}} 

\newcommand{\hXij}{\widehat{x}_{ij}}

\newcommand{\Xdij}{x^\bullet_{ij}}

\newcommand{\Ydii}{y^\bullet_{ii}}
\newcommand{\Ydjj}{y^\bullet_{jj}}
\newcommand{\Xddij}{x^{\bullet\bullet}_{ij}}
\newcommand{\Xsdij}{m^\bullet_{ij}}
\newcommand{\Xsddij}{m^{\bullet\bullet}_{ij}}

\newcommand{\shurdij}{\shurij^\bullet}
\newcommand{\Delij}{\del_{ij}}

\newcommand{\Xkij}{x^k_{ij}}

\newcommand{\Ykij}{y^k_{ij}}
\newcommand{\Ykii}{y^k_{ii}}
\newcommand{\Ykjj}{y^k_{jj}}
\newcommand{\shurkij}{\shurij^k}
\newcommand{\Xkpij}{x^{k+1}_{ij}}
\newcommand{\Xskij}{m^k_{ij}}
\newcommand{\Xkji}{x^k_{ji}}

\newcommand{\Xkn}{\X^{k_n}}

\newcommand{\Xknij}{\Xij^{k_n}}
\newcommand{\Xknbpij}{\Xij^{k_n+1}}

\newcommand{\Xknspij}{\Xij^{k_{n+1}}}

\newcommand{\Yknij}{\Yij^{k_n}}

\newcommand{\Yknii}{\Yii^{k_n}}
\newcommand{\Yknjj}{\Yjj^{k_n}}

\newcommand{\pllknij}{\phi_{\X^{k_n},ij}}

\newcommand{\pllcknij}{c_{\X^{k_n},ij}}

\newcommand{\Xsknij}{\Xsij^{k_n}}
\newcommand{\Xsknbmij}{\Xsij^{k_n-1}}

\newcommand{\shurknij}{\shurij^{k_n}}

\newcommand{\Indkn}{\Ind_{k^n}}
\newcommand{\Indmapkn}{\Ind_{k^n+1}}

\newcommand{\ei}{\mb{e}_i}
\newcommand{\ej}{\mb{e}_j}
\newcommand{\Mati}{\mb{x}_{[i]}}
\newcommand{\Yi}{\mb{y}_{[i]}}
\newcommand{\Yj}{\mb{y}_{[j]}}

\newcommand{\tr}{\textup{tr}}
\newcommand{\diag}{\textup{diag}}

\newcommand{\Ind}{\mbb{I}}
\newcommand{\sgn}{\textup{sgn}}
\newcommand{\kronp}{\otimes}

\newcommand{\ith}{i^{\textup{th}}}
\newcommand{\ijth}{ij^{\textup{th}}}
\newcommand{\PLL}{\mc{L}}

\newcommand{\pllij}{\phi_{\Xo,ij}}
\newcommand{\pllii}{\phi_{\Xo,ii}}

\newcommand{\pllcij}{c_{\Xo,ij}}

\newcommand{\pllKij}{\phi_{\Xk,ij}}

\newcommand{\pllcKij}{c_{\Xk,ij}}

\newcommand{\SLo}{\mc{S}_{l_0}}

\newcommand{\Uijval}{\begin{bmatrix} \ei & \ej \end{bmatrix}}

\newcommand{\Zc}{\mc{Z}^C}
\newcommand{\eps}{\epsilon}

\newcommand{\Ord}{\mc{O}}

\newcommand{\As}{\mc{S}_{A}}

\newcommand{\Map}{\mc{A}}

\newcommand{\bq}{\bar{q}}

\newcommand{\Zoij}{x^+_{0,ij}}
\newcommand{\Df}{\mc{D}}
\newcommand{\Dfs}{\Df^\infty}

\newcommand{\Indd}{\Ind^\bullet}
\newcommand{\Inddd}{\Ind^{\bullet\bullet}}
\newcommand{\Inds}{\Delta\Ind^\infty}

\newcommand{\PLLk}{\mc{L}_{k}}
\newcommand{\PLLkp}{\mc{L}_{k+1}}

\newcommand{\Fix}{\mc{F}}
\newcommand{\Fixij}{\mc{F}_{ij}}

\newcommand{\sqoij}{\square_{0,ij}}

\newcommand{\LL}{l_\lam}
\newcommand{\Nh}{\mc{U}}

\newcommand{\Zs}{\mc{Z}}
\newcommand{\Zcs}{\mc{Z}^c}

\newcommand{\Res}{\textup{R}_\lam}
\newcommand{\Sz}{\textup{S}_{\mc{Z}^c}}
\newcommand{\Szc}{\textup{S}_{\mc{Z}}}

\newcommand{\epsp}{\eps^\prime}
\newcommand{\cij}{c_{ij}}

\newcommand{\Mset}{\mc{M}}

\newcommand{\mukl}{\textup{KL}}

\newcommand{\hmukl}{\widehat{\textup{KL}}}

\newcommand{\Seto}{\mc{X}}

\begin{document}

\title{$l_0$ Sparse Inverse Covariance Estimation}

\author{Goran~Marjanovic,~\IEEEmembership{Member,~IEEE,}
        and~Alfred~O.~Hero~III,~\IEEEmembership{Fellow,~IEEE}
\thanks{Goran Marjanovic is with the School of Electrical Engineering, University of New South Wales, Sydney, Australia e-mail: g.marjanovic@unsw.edu.au}
\thanks{Alfred O. Hero III is with the Department
of Electrical and Computer Science, University of Michigan, Ann Arbor,
MI, USA e-mail: hero@eecs.umich.edu}
\thanks{This research was partially supported by AFSOR grant FA9550-13-1-0043.}
}


\maketitle

\begin{abstract}
Recently, there has been focus on penalized log-likelihood covariance estimation for sparse inverse covariance (precision) matrices. The penalty is responsible for inducing sparsity, and a very common choice is the convex $l_1$ norm. However, the best estimator performance is not always achieved with this penalty. The most natural sparsity promoting ``norm'' is the non-convex $l_0$ penalty but its lack of convexity has deterred its use in sparse maximum likelihood estimation. In this paper we consider non-convex $l_0$ penalized log-likelihood inverse covariance estimation and present a novel cyclic descent algorithm for its optimization. Convergence to a local minimizer is proved, which is highly non-trivial, and we demonstrate via simulations the reduced bias and superior quality of the $l_0$ penalty as compared to the $l_1$ penalty.
\end{abstract}


\section{Introduction}

Graphical models have a long history \cite{Dempster72,Whittaker90,Lauritzen96} and provide a systematic way of analyzing dependencies in high dimensional data. The structure of the graph identifies meaningful interactions among the data variables. When the data is Gaussian with mean $\mb{0}_{p\times 1}$ and covariance $\Sig_{p\times p}$, the graphical model is an undirected graph specified by the non-zeros in the precision (inverse covariance) matrix $\Om=\Sig^{-1}$. In this Gaussian case the graph captures conditional dependency (Markovian) properties of the variables: the absence of an edge between nodes $i$ and $j$, $i\neq j$, in the graph reflects conditional independence of variables $i$ and $j$ given the other variables. Letting $\Omij$ denote the $ij$-th component of $\Om$, this in turn corresponds to having $\Omij=0$, \cite{Dempster72,Whittaker90,Lauritzen96}.

Following the parsimony principle, the estimation objective is to choose the simplest model, i.e., the sparsest graph that adequately explains the data. The sparsity requirement improves the interpretability of the model and reduces over-fitting. In order to estimate a sparse $\Om$, much attention has been given to minimizing a sparsity \textbf{P}enalized \textbf{L}og-\textbf{L}ikelihood (\textbf{PLL}) objective function. The log-likelihood promotes goodness-of-fit of the estimator while the penalty promotes many of its entries to become zero. 

Even though the $l_0$ ``norm'' \footnote{The $l_q$ function is not a norm for $q<1$.} is the natural sparsity promoting penalty, the $l_1$ norm has become its dominant replacement. The primary justification is the convexity of the $l_1$ penalty and this has resulted in its widespread use in sparse linear regression \cite{Tibshirani96}. As the $l_1$-PLL objective function is convex, convex optimization approaches can be applied to obtain sparse penalized \textbf{M}aximum-\textbf{L}ikelihood (\textbf{ML}) estimators. As a result, there has been extensive research in the development of efficient methods for solving the $l_1$-PLL problem. Examples include \cite{FHT08,BGd08,RBLZ08,SMG10,SR10,YY10,Yuan09,HSDR11,OONR12,dBE08,DGK08,LT10}, and an overview is given \cite{RG11,SNW11}. These methods range from cyclic descent type algorithms \cite{dBE08,FHT08,SR10,RBLZ08}, to alternating linearization algorithms \cite{SMG10,YY10,Yuan09}, and projected sub-gradient methods \cite{DGK08}. Newton-type methods that incorporate cyclic descent, conjugate gradient as well as iterative shrinkage methods \cite{BT209}, are considered in \cite{HSDR11,OONR12}. 

Despite the high popularity of the $l_1$ norm in sparsity penalized ML estimation problems, it has certain drawbacks. One drawback is that $l_1$ penalization induces shrinkage of the parameter estimates, which introduces negative biases \cite{FFW09,LF09,MHT10,FL01}. Another drawback is that for very sparse problems $l_1$-PLL does not produce sufficiently sparse estimates \cite{FFW09,MHT10,Friedman08,MS12}, resulting in the recovery of less parsimonious models. Hence, it is natural to ask the question: can the $l_0$ penalized estimator of inverse covariance provide improvement over the $l_1$ penalized estimator? The $l_0$ penalty has been considered in other sparsity penalized problem formulations, for example, in sparse linear regression \cite{SS12,BYD07,BD208,Nikolova13,ZDL13,DZ13,MUH15}, sparse signal recovery \cite{Nikolova13}, PCA and low rank matrix completion \cite{MHT10,US11,USM15}. The $l_0$ penalty induces maximum sparsity and would be expected to have superior prediction accuracy relative to $l_1$ penalized PLL, especially for very sparse $\Om$.

In this paper we develop an algorithm for solving the non-convex $l_0$-PLL problem for inverse covariance estimation. We propose a novel \tb{C}yclic \tb{D}escent (\textbf{CD}) algorithm to implement the optimization. We prove convergence of the algorithm to a local minimizer of the $l_0$-PLL objective function. 

CD algorithms developed for optimizing the $l_1$-PLL objective function are proposed in \cite{FHT08,SR10,RBLZ08,BGd08,SMG10,HSDR11}. The GLASSO method in \cite{FHT08} and its variant in \cite{BGd08} are block-type CD procedures, which are derived using duality arguments and convergence analysis is performed using convexity arguments. The method in \cite{RBLZ08} applies the CD procedure to the elements of the Cholesky decomposition of each iterate. The SINCO method in \cite{SR10} is a greedy-type algorithm derived using an equivalent reformulation of the $l_1$-PLL problem by exploiting the piecewise linearity of the $l_1$ penalty. The ALM algorithm in \cite{SMG10} uses linearization to find solutions of the objective function surrogates, which are updated in an alternating fashion. These iterates eventually converge to a single solution. The QUIC algorithm in \cite{HSDR11} is a quasi-Newton type method, which applies an efficient CD procedure on a second order approximation of the $l_1$-PLL objective function. Inexact line search is then used to achieve descent. QUIC is a special case of the Newton-type methods proposed in \cite{OONR12}. To minimize the second order approximation, \cite{OONR12} also considers the nonlinear conjugate gradient method and the FISTA algorithm from \cite{BT209}. The latter is a Majorization-Minimization or a proximal-type method. A monotone version of FISTA, called M-FISTA, from \cite{BT309} can also be considered to improve stability. 

Due to non-linearity and non-convexity of the $l_0$-PLL objective function, we cannot exploit any of the above ideas to derive $l_0$-PLL algorithms and analyze their convergence. Alternating linearization procedures are extremely hard to analyse in the non-convex setting, and could result in unstable algorithms if applied blindly. Furthermore, we cannot exploit second order approximations because the inexact line-search techniques used for convex criteria cannot be easily modified to guarantee descent for the non-convex $l_0$-PLL criterion. So, instead of attempting to modify existing $l_1$ based methods, we have to rely on direct arguments, which make our algorithm fundamentally different. Additionally, the proposed method uses coordinate-by-coordinate optimization and, hence, is fundamentally different from those in \cite{MH14,GMThesis13,MS11} that utilize a block-type CD procedure.

The remainder of the paper is organized as follows. Section \ref{notation section} gives necessary notation, while Section \ref{problem section} introduces the $l_0$-PLL problem. The CD algorithm is derived in Section \ref{algorithm section}, and the convergence analysis is provided in Section \ref{convergence section}. Finally, Section \ref{simulations section} contains simulation results and Section \ref{conclusion section} has the conclusion.

\section{Notation} \label{notation section}

For a square matrix $\Mat_{p\times p}=[\Mij]$, the $\ijth$ element is denoted by $[\Mat]_{ij}=\Mij$, and the $\ith$ column vector is denoted by $\Mati$. We write $\det(\Mat)$ for the determinant of $\Mat$, and $\tr(\Mat)$ for the trace of $\Mat$. The notation $\diag(\Mat)$ denotes a vector containing the diagonal elements of $\Mat$.
We write $\Mat\succ 0$ and $\Mat\succeq 0$ to indicate that $\Mat$ is positive definite and positive semi-definite respectively. $\Ind(\cdot)$ denotes the indicator function, equaling $1$ if the argument is logically true, and $0$ otherwise. $\sgn(\cdot)$ denotes the sign function. $\ei$ is a unit vector with a $1$ in the $\ith$ entry and $0$ in all other entries. Using this unit vector definition, we also define the $p\times 2$ matrix: 
\begin{equation}
\Uij=\Uijval,  \label{Uij and Vij def}
\end{equation}
$(\cdot)^T$ denotes the transpose operator, and $\|\cdot\|_F$ denotes the Frobenius (matrix) norm. $\kronp$ denotes the Kronecker product. Lastly, $\{x_k\}_k$ denotes a sequence $x_1$, $x_2$, .... The sequence $\{x_{k_n}\}_n$ denotes a subsequence of $\{x_k\}_k$, where $n=1,2,\dots$, i.e., $\{k_n\}\subseteq\{k\}$, and $k_n\leq k_{n+1}$ for all $n$.

\section{The $l_0$ Penalized Log-Likelihood Problem} \label{problem section}

In this section we introduce the $l_0$-PLL problem formulation in the multivariate Gaussian setting. Define the $l_0$ ``norm'' for any $\Mat_{p\times p}$:
\begin{equation}
\|\Mat\|_0=\sum_{i=1}^p\sum_{j=1}^p\Ind(\Mij\neq 0). \label{L0 norm}
\end{equation}
Denote the sample covariance matrix by $\Samp_{p\times p}=[\Sampij]$ which, by definition, is symmetric and positive semi-definite. We assume that $\Samp$ is constructed from $n$ independent samples drawn from a $p$-variate Gaussian distribution with mean $\mb{0}_{p\times 1}$ and covariance $\Sig_{p\times p}$. We additionally assume that $\Sampii>0$ for all $i=1,\dots,p$. Recalling that $\Om=\Sig^{-1}\succ 0$, the aim is to estimate a sparsified $\Om$ by minimizing (at least locally) the following non-convex $l_0$-PLL objective function:
\begin{equation}
\PLL(\X)=-\log\det(\X)+\tr(\Samp\X)+\lam\|\X\|_0, \label{L0-PLL}
\end{equation}
over the space of symmetric and positive definite matrices $\X$, where $\lam>0$ is a tuning parameter. We recall that the $l_1$-PLL objective function is obtained by replacing the $l_0$ penalty in (\ref{L0-PLL}) by the $l_1$ norm of the matrix entries, i.e., by:
\begin{equation}
\|\X\|_1=\sum_{i=1}^p\sum_{j=1}^p\left|\Mij\right| \label{L1 norm}
\end{equation}
see \cite{FHT08,BGd08,RBLZ08,SMG10,SR10,YY10,Yuan09,HSDR11,OONR12,dBE08,DGK08,LT10}.

An important question is whether the solution of the $l_1$-PLL problem, with some tuning parameter $\mu>0$, is also a minimizer of (\ref{L0-PLL}). The answer is no, as given in the following theorem:
\newtheorem{L0 vs L1 solutions theorem}{Theorem}
\begin{L0 vs L1 solutions theorem} \label{L0 vs L1 solutions thm}
\textup{Suppose $\hOmLon(\mu)$ is a global minimizer of the $l_1$-PLL objective function with tuning parameter $\mu>0$. Denote the set of all local minimizers of (\ref{L0-PLL}) by $\SLo(\lam)$. Then $\hOmLon(\mu)\notin\SLo(\lam)$ for any $\mu>0$.
}
\end{L0 vs L1 solutions theorem}

\textit{Proof.} See Appendix B.

\vspace{1mm}

Since all global minimizers are also local minimizers, Theorem \ref{L0 vs L1 solutions thm} implies that any solution of the $l_1$-PLL problem will not be a global minimizer of (\ref{L0-PLL}). As a result, this theorem motivates a different approach to minimizing (\ref{L0-PLL}).

\section{Algorithm Development} \label{algorithm section}

In this section we derive a \textbf{C}oordinate \textbf{D}escent (\textbf{CD}) algorithm for finding local minima of (\ref{L0-PLL}). 

The basic concept of the algorithm is to fix all entries except for one selected entry of the current (symmetric) iterate $\Xo\succ 0$. $\PLL(\cdot)$ is then minimized with respect to (w.r.t.) the selected entry. Once the new value of this entry is calculated, $\Xo$ is updated and $\PLL(\cdot)$ is minimized w.r.t. the next selected entry. The update equation is:
\begin{equation}
\Zij(\Xij)=\Xo+
\begin{cases}
(\Xii-\Xoii)\ei\ei^T & \textup{if }i=j \\
(\Xij-\Xoij)\Uij\Vij^T & \textup{otherwise}, \label{X(xij) function}
\end{cases}
\end{equation}
where $\Uij$ is defined in (\ref{Uij and Vij def}). For what follows we define: 
\begin{equation}
\Yo=\Xo^{-1} \textup{ and }\del(\Xij)=\Xij-\Xoij,    \label{Yo and del}
\end{equation}
as well as:
\begin{align}
\pllij(\Xij)&=-\log\det(\Zij(\Xij))+\Sampij\Xij  \nonumber\\
&\hspace{10mm}+\left\{\Sampij\Xij+ 2\lam\Ind(\Xij\neq 0)\right\}\Ind(i\neq j),  \label{pll for any ij}
\end{align}
for any $i,j$. We will also rely on the standard determinant and matrix inverse identities given in Appendix A. 

\subsection{Element-wise Minimizers of $\PLL(\cdot)$ when $i=j$}

The minimizers of $\PLL(\Zii(x))$ are given by:
\begin{align}
&\arg\min_{x}\ \pllii(x)  \nonumber\\
=&\arg\min_{x}\ -\log\det(\Zii(x))+\Sampii x \nonumber\\
=&\arg\min_{x}\ \PLL(\Zii(x)), \nonumber
\end{align}
where $\pllii(\cdot)$ is defined in (\ref{pll for any ij}). Noting that $\pllii(\cdot)$ is differentiable, the minimizers are given by solving the equation:
\begin{equation}
\pllii^\prime(x)=-[\Zii(x)^{-1}]_{ii}+\Sampii =0. \label{pll der for ii}
\end{equation}
We substitute $\del=\del(x)$ and $\X=\Zii(x)$ in the matrix inverse identity (\ref{inv of perturb ii}) to obtain:
\begin{equation}
[\Zii(x)^{-1}]_{ii}=\frac{\Yoii}{1+\del(x)\Yoii}. \label{Xii inv result}
\end{equation}
So, substituting (\ref{Xii inv result}) in (\ref{pll der for ii}) and solving for $\Xii$, the (unique) minimizer is given by:
\begin{equation}
\Xsii=\Xoii+\frac{\Yoii-\Sampii}{\Yoii\Sampii}. \label{mii value}
\end{equation}
We lastly need to check that $\Zii(\Xsii)\succ 0$, i.e., is invertible. By observing (\ref{det of perturb ii}) or (\ref{inv of perturb ii}), this requires that $1+\del(\Xsii)\Yoii>0$, which can easily be confirmed.

\subsection{Element-wise Minimizers of $\PLL(\cdot)$ when $i\neq j$} \label{mij when i neq j}

The minimizers of $\PLL(\Zij(x))$ are given by:
\begin{align}
&\arg\min_{x}\ \pllij(x) \nonumber\\
=&\arg\min_{x}\ -\log\det(\Zij(x))+2\Sampij x + 2\lam\Ind(x\neq 0) \nonumber\\
=&\arg\min_{x}\ \PLL(\Zij(x)), \nonumber
\end{align}
where $\pllij(\cdot)$ is again defined in (\ref{pll for any ij}). In this case, $\pllij(\cdot)$ has a single discontinuity at $x=0$ but only if $0$ is in the domain of $\pllij(\cdot)$, i.e., if $\det(\Zij(0))>0$. Otherwise, $\pllij(\cdot)$ would be continuous everywhere. The continuous (and differentiable) part of $\pllij(\cdot)$ is given by:
\begin{equation}
\pllcij(\Xij)=-\log\det(\Zij(\Xij))+2\Sampij\Xij + 2\lam. \label{pllc fun}
\end{equation}
First consider the case that $\det(\Zij(0))>0$, in which case we can equivalently express $\pllij(\cdot)$ as:
\begin{align}
\pllij(\Xij)&=\pllcij(\Xij)\Ind(\Xij\neq 0) \nonumber\\
&\hspace{15mm} + (\pllcij(\Xij)-2\lam)\Ind(\Xij=0). \label{pll fun}
\end{align}
Now we see that the minimizers of $\pllij(\cdot)$ are the minimizers of $\pllcij(\cdot)$ or $\Xij=0$. Since $\pllcij(\cdot)$ is strictly convex, it has a unique minimizer obtained as the solution to:
\begin{equation}
\pllcij^\prime(x)=-2[\Zij(x)^{-1}]_{ij}+2\Sampij=0. \label{pllc der for ij}
\end{equation}
Substituting $\del=\del(x)$ and $\X=\Zij(x)$ into the matrix inverse identity (\ref{inv of perturb ij}), we obtain:
\begin{equation}
[\Zij(x)^{-1}]_{ij}=\frac{-\shuroij\del(x)+\Yoij}{-\shuroij\del(x)^2+2\Yoij\del(x)+1}, \label{Xij inv result}
\end{equation}
where 
\begin{equation}
\shuroij=\shurij(\Yo)>0, \label{shur value}
\end{equation}
and $\shurij(\cdot)$ is given by (\ref{shur quantity}). Substituting (\ref{Xij inv result}) into (\ref{pllc der for ij}) and solving for $x$, the (unique) minimizer is:
\begin{equation}
\Xsij=\Xoij+\frac{\Yoij}{\shuroij}, \label{mij value 1}
\end{equation}
when $\Sampij=0$. $\Zij(\Xsij)\succ 0$ since by (\ref{det of perturb ij}) $-\shuroij\del(\Xsij)^2+2\Yoij\del(\Xsij)+1>0$.

When $\Sampij\neq 0$, by substituting (\ref{Xij inv result}) into (\ref{pllc der for ij}), (\ref{pllc der for ij}) is equivalent to:
\begin{equation}
\shuroij\Sampij\del(\Xij)^2-(\shuroij+2\Yoij\Sampij)\del(\Xij)+(\Yoij-\Sampij)=0. \nonumber 
\end{equation}
The discriminant of the above quadratic equation is: $\shuroij^2+4\Sampij\Yoii\Yojj>0$, and so, there are two solutions. However, only one of these, given by: 
\begin{equation}
\hspace{-0.5mm} \Xsij=\Xoij+\frac{\Yoij}{\shuroij}+\frac{\shuroij-\sqrt{\shuroij^2+4\Sampij^2\Yoii\Yojj}}{2\shuroij\Sampij} \label{mij value 2}
\end{equation}
yields $-\shuroij\del(\Xsij)^2+2\Yoij\del(\Xsij)+1>0$, i.e., $\Zij(\Xsij)\succ 0$. Note that, from L'Hopital's rule, (\ref{mij value 2}) approaches (\ref{mij value 1}) as $\Sampij\to 0$.

Lastly, $\det(\Zij(0))\leq 0$ implies that the (unique) minimizer of $\pllij(\cdot)$ in (\ref{pll for any ij}) is equal to $\Xsij$.

The above results are summarized in the following theorem: 
 
\newtheorem{mij is not zero theorem}[L0 vs L1 solutions theorem]{Theorem}
\begin{mij is not zero theorem} \label{mij is not zero thm}
\textup{
When $i\neq j$, the minimizers $\hXij$ of $\pllij(\cdot)$ in (\ref{pll for any ij}) satisfy:\\
\noindent $\bullet$ when, $\det(\Zij(0))\leq 0$:
\begin{equation}
\hXij=\Xsij, \label{mij expression 1}
\end{equation}
\noindent $\bullet$ when, $\det(\Zij(0))>0$:
\begin{equation}
\hXij=
\begin{cases}
0 & \textup{ if }\pllij(0)<\pllij(\Xsij) \\
\{0,\Xsij\} & \textup{ if }\pllij(0)=\pllij(\Xsij) \\
\Xsij & \textup{ if }\pllij(0)>\pllij(\Xsij), \label{mij expression 2}
\end{cases}
\end{equation}
where $\Xsij=\Xsij(\Xij)$ is given by (\ref{mij value 1}) when $\Sampij=0$, and is given by (\ref{mij value 2}) otherwise, and $\pllij(\cdot)$ is given by (\ref{pll fun}).
}
\end{mij is not zero theorem}

\subsection{Dealing with $\det(\X(0))$, $\pllij(0)$ and $\pllij(\Xsij)$}

Computing (\ref{mij expression 1}) and (\ref{mij expression 2}) requires two operations:
\begin{itemize}
\item[(a)] comparing $0$ to $\det(\Zij(0))$
\item[(b)] comparing $\pllij(0)$ to $\pllij(\Xsij)$
\end{itemize}
Even though all the mentioned quantities contain $\det(\Zij(\cdot))$, (a) and (b) must be done efficiently without explicitly calculating the determinant. 

For (a), we substitute $\del=-\Xoij$ and $\X=\Xo$ into the determinant identity (\ref{det of perturb ij}) and, since $\det(\Xo)>0$,
\begin{align}
-\shuroij\Xoij^2 - 2\Yoij\Xoij + 1>0. \label{detXo>0}
\end{align}

For (b), we again substitute $\del=-\Xoij$ and $\X=\Xo$ into (\ref{det of perturb ij}) to obtain an expression for $\pllij(0)$, i.e.,
\begin{align}
\pllij(0)&=-\log\det(\Zij(0)) \nonumber\\
&=-\log\det(\Xo) \nonumber\\
&\hspace{4mm}-\log\{-\shuroij\Xoij^2-2\Yoij\Xoij+1\}. \label{pll(0) expression}
\end{align}
Then, substituting $\del=\Xsij-\Xoij$ and $\X=\Xo$ in (\ref{det of perturb ij}) we obtain an expression for $\pllij(\Xsij)$
\begin{align}
\pllij(\Xsij)&=\pllcij(\Xsij) \nonumber\\[2mm]
&=-\log\det(\Xo) \nonumber\\[2mm]
&\hspace{4mm}-\log\left(-\shuroij\del(\Xsij)^2+2\Yoij\del(\Xsij)+1\right) \nonumber\\[2mm]
&\hspace{4mm}+2\Sampij\Xsij+2\lam. \label{pllc(Xsij) expression}
\end{align}
When comparing $\pllij(0)$ to $\pllij(\Xsij)$, expressions (\ref{pll(0) expression}) and (\ref{pllc(Xsij) expression}) lead to an expression that is minimized without the need for explicit calculation of any matrix determinants.

\subsection{Updating $\Yo=\Xo^{-1}$} \label{updating the inverse subsection}

Since $\Yo$ is needed to compute the entry update (\ref{mii value}), (\ref{mij value 1}) and (\ref{mij value 2}), $\Yo$ needs to be updated as well. An efficient way to do this is to use the matrix inverse identities (\ref{inv of perturb ii}) and (\ref{inv of perturb ij}) with substitutions $\del=\del(\Xsij)$ and $\X=\Xo$. 

After every off-diagonal entry update the proposed CD algorithm needs to compute a new matrix inverse $\Yo$, which requires $\Ord(p^2)$ multiplications. As a result, there are order $\frac{1}{2}p^2\times\Ord(p^2)$ multiplications for each matrix sweep. Now, note that if:
\begin{align}
\Xoij=0 \textup{ and } \pllij(0)\leq\pllij(\Xsij), \label{the active set cond}
\end{align}
then there is no change in $\Xo$, and hence $\Yo$ would not need to be updated. In practice, the sparser the problem we are dealing with the larger the set of entries that satisfy (\ref{the active set cond}) becomes, resulting in a smaller (``active'') set of entries for which $\Yo$ is updated. Thus, the $\frac{1}{2}p^2$ factor in the inverse updating can in practice be reduced to something close to just half the number of off-diagonal non-zeros in $\Xo$; a much smaller number.

\newtheorem{order of products remark}{Remark}
\begin{order of products remark} \label{order of products rmk}
\textup{
To make sure that the size of the ``active'' set is small the CD algorithm should be initialized with a very sparse matrix, e.g., a diagonal matrix.   
}
\end{order of products remark}

\subsection{Coordinate Descent (CD) Algorithm for the $l_0$ Penalized Log-Likelihood ($l_0$-PLL) Problem}

Here we state the CD algorithm for minimizing (\ref{L0-PLL}). 

\subsubsection{Initialization} \label{alg init}

\vspace{1mm}

From \cite[Theorem 3]{MH14} we know that a necessary condition for existence of a solution to (\ref{L0-PLL}) is $1/\Sampii>0$. In order to guarantee a small active set, following Remark \ref{order of products rmk}, we initialize the CD algorithm with $1/\Sampii$ for every $i=1,\dots,p$. 

\subsubsection{Updating the Entries}

\vspace{1mm}

Note that only the diagonal entries and only half of the off-diagonal entries need to be updated. Denote the set of indices of all these entries by $\As$, which is easily computed off-line. For very large and very sparse problems the CD algorithm can be sped-up by only updating the non-zero components after a sufficiently large number of matrix sweeps. Updating only a subset of entries per matrix sweep is used for CD algorithm speed-ups for minimizing the convex $l_1$-PLL objective function \cite{HSDR11,OONR12}. 



\noindent\rule{\columnwidth}{0.052cm}\\*[-0.05cm]
\textbf{The Coordinate Descent (CD) Algorithm}\\*[-0.25cm]
\rule{\columnwidth}{0.03cm}\\*[-0.5cm]
\begin{enumerate}
\item[(1)] Suppose $\Xk=[\Xkij]$ and $\Yk=(\Xk)^{-1}=[\Ykij]$ are the current iterates (symmetric).
\vspace{1mm}
\item[(2)] Let $\Xo=\Xk$ and $\Yo=\Yk$, and for each $(i,j)\in\As$, repeat (i) to (vi):\\[2mm]
(i) $\Xskij=\Xskij(\Xkij)$ is set according to: 
\begin{itemize}
\item[$\bullet$] (\ref{mii value}) if $i=j$.
\item[$\bullet$] (\ref{mij value 1}) if $i\neq j$ and $\Sampij=0$.
\item[$\bullet$] (\ref{mij value 2}) if $i\neq j$ and $\Sampij\neq 0$.
\end{itemize}
\vspace{1mm}
(ii) If $-\shurkij(\Xkij)^2-2\Ykij\Xkij+1>0$ \underline{and} $i\neq j$, compute:
\small
\begin{equation}
\hspace{-3mm}\Map(\Xkij)=
\begin{cases}
0 & \hspace{-1mm}\textup{if } \pllKij(0)<\pllKij(\Xskij) \\[2mm]
\Xskij\hspace{0.5mm}\Ind(\Xkij\neq 0) & \hspace{-1mm}\textup{if } \pllKij(0)=\pllKij(\Xskij) \\[2mm]
\Xskij & \hspace{-1mm}\textup{if } \pllKij(0)>\pllKij(\Xskij) {\normalsize\label{algorithm map 1}}
\end{cases}
\end{equation}
\normalsize
where $\shurkij=\shurij(\Yk)$, and $\shurij(\cdot)$ is given by (\ref{shur quantity}). \\[2mm]
(iii) If $-\shurkij(\Xkij)^2-2\Ykij\Xkij+1\leq 0$ \underline{or} $i=j$, compute:
\begin{equation}
\Map(\Xkij)=\Xskij \label{algorithm map 2}
\end{equation}
(iv) Update $\Xkij$ (and $\Xkji$ if $i\neq j$) with:
\begin{equation}
\Xkpij=\Map(\Xkij) \label{xij update}
\end{equation}
(v) Denote the matrix with the updated $\Xkij$ by $\Xkp$. Then, calculate $\Ykp=(\Xkp)^{-1}$ using the Sherman Morrison Woodbury formula: Let 
\begin{align}
\del=\Xkpij-\Xkij. \nonumber
\end{align}
If $\del\neq 0$, then:

\noindent\hspace{2mm} $\bullet$ for $i=j$:\\[-6mm]
\small
\begin{align}
\Ykp=\Yk-\del\frac{\Yi^k\Yi^{k,T}}{1+\del\Ykii} \nonumber 
\end{align}
\normalsize
\hspace{2mm} $\bullet$ for $i\neq j$:\\[-6mm]
\small
\begin{align}
\hspace{-3mm}\Ykp=\Yk-\del\frac{\begin{bmatrix} \Yi^k &\hspace{-2mm} \Yj^k \end{bmatrix}
\begin{bmatrix} 1+\del\Ykij & -\del\Ykjj \\ -\del\Ykii & 1+\del\Ykij \end{bmatrix}
\begin{bmatrix} \Yj^{k,T} \\[1mm] \Yi^{k,T} \end{bmatrix}}
{-\shurkij\del^2+2\Ykij\del+1} \nonumber
\end{align}
\normalsize
(vi) Increment the counter $k$ by $1$.
\vspace{1mm}
\item[(3)] Go to (1).
\vspace{-2mm}
\end{enumerate}
\rule{\columnwidth}{0.052cm}

\newtheorem{alg remark}[order of products remark]{Remark}
\begin{alg remark} \label{alg rmk}
\textup{The map $\Map(\Xoij)$ depends on $\Xo$ in step (2) of the algorithm as well as indices $ij$. It is given by the element-wise minimizer $\hXij$ in (\ref{mij expression 1}) and (\ref{mij expression 2}). Since in (\ref{mij expression 2}) we see that there are two minimizers $0$ and $\Xsij$, we have set $\Map(\Xoij)$ to $0$ when the current value is $0$, and to $\Xsij$ otherwise. The motivation for this choice is Theorem \ref{Map is fixed point thm} in the next section.
}
\end{alg remark}


\section{Convergence Analysis} \label{convergence section}

Convergence of CD methods for sparse and general problems have been previously analysed \cite{MFH11,Tseng01,LY08,Bertsekas99,WGS,dBE08,FHT08,SR10,RBLZ08}. The analysis in \cite{LY08,Bertsekas99,WGS} holds only for convex functions, and is not applicable. Convergence has been proved in \cite{Tseng01} under weaker convexity assumptions. However, these assumptions do not hold for the $l_0$-PLL problem. Lastly, the global convergence theorem in \cite{LY08,Luenberger73} fails because $\PLL(\cdot)$ is not continuous, furthermore the lack of differentiability prevents us from using any analysis in \cite{MFH11}.

In the following convergence analysis we firstly use the algorithm map $\Map(\cdot)$ to show that the fixed points of the algorithm are strict local minimizers. Then, under two necessary conditions it is subsequently shown that the whole sequence converges to a single local minimizer.

\newtheorem{sequence remark}[order of products remark]{Remark}
\begin{sequence remark} \label{sequence rmk}
\textup{
The statement $\Xkij\to\Xdij$ as $k\to\infty$ applies to the fixed $ij$-th entry of $\Xk$. Due to the cyclic nature of the CD algorithm, this means that $k$ is a function of $(i,j)$, i.e., $k=k(i,j)=i+(j-1)p + p^2r$ and $r=0,1,2,\dots$. For example, if the size of $\Xk$ is $p=4$ and we focus on entry $(3,2)$, then $k=7,23,39,\dots,\infty$ corresponds to the iterations where this entry is updated. In order to simplify notation the iteration counter in $\Xkij$ will simply be denoted by $k$, noting that we actually mean $k(i,j)$. Since the fixed $ij$-th entry in statement $\Xkij\to\Xdij$ is arbitrary, the statement is therefore equivalent to the statement $\Xk\to\Xd$. 
}
\end{sequence remark}     

The set of fixed points of the algorithm is defined as:
\begin{equation}
\Fix=\bigcap_{ij}\Fixij, \textup{ where }\Fixij=\left\{\X\succ 0:\Xij=\Map(\Xij)\right\}. \label{F set}
\end{equation}

where $\Fixij$ is the set of positive definite matrices that satisfy the fixed point equation $\Xij=\Map(\Xij)$. The definition of $\Map(\cdot)$ in (\ref{algorithm map 1}) asserts that $\X^k$ converges to a fixed point $\X^\bullet$ of $\Map(\cdot)$: 
\newtheorem{Map gives fixed point theorem}[L0 vs L1 solutions theorem]{Theorem}
\begin{Map gives fixed point theorem} \label{Map is fixed point thm}
\textup{
If $\Xkij\to\Xdij$ as $k\to\infty$, then $\Xdij=\Map(\Xdij)$, i.e., $\Xd\in\Fix$.
}
\end{Map gives fixed point theorem}

\textit{Proof.} See Appendix B. 

\vspace{2mm}

The following theorem establishes that the fixed points are isolated points and hence strict local minimizers of (\ref{L0-PLL}): 

\newtheorem{F is local min theorem}[L0 vs L1 solutions theorem]{Theorem}
\begin{F is local min theorem} \label{F is local min thm}
\textup{$\X\in\Fix$ is a strict local minimizer of $\PLL(\cdot)$. Specifically, there exists $\eps>0$ such that for any symmetric $\Del=[\Delij]$ satisfying $0<\|\Del\|_F<\eps$:
\begin{equation}
\PLL(\X)<\PLL(\X+\Del). \label{PLL inequality}
\end{equation}
}
\end{F is local min theorem}

\textit{Proof.} See Appendix B. 

\vspace{2mm}

Theorems \ref{Map is fixed point thm} and \ref{F is local min thm} imply that a convergent algorithm must converge to a local minimizer. 

Next, consider the following two assumptions:
 
\newtheorem*{bounded sequence assumption}{(A1)}
\begin{bounded sequence assumption} \label{bounded sequence ass} 
\textup{Assume there exists a $K>0$ and $\alpha\in(0,\infty)$ such that $\Xk\preceq\alpha\Id$ for all $k>K$.
} 
\end{bounded sequence assumption}

\newtheorem*{xkn assumption}{(A2)}
\begin{xkn assumption} \label{xkn ass} 
\textup{For any subsequence $\{\Xknij\}_n$ such that $\lim_{n\to\infty} \Xknij\in\Seto_{\Xo}=\left\{\Xoij: i\neq j,\ \pllij(0)=\pllij(\Xsij)\right\}$, assume:
\begin{itemize}
\vspace{2mm}
\item[(a)] $\Xknij=0$ $\forall n>N$ implies $\Xknbpij=0$ $\forall n>N$,
\vspace{2mm}
\item[(b)] $\Xknij\neq 0$ $\forall n>N$ implies $\Xknbpij\neq 0$ $\forall n>N$,
\vspace{2mm}
\end{itemize} 
where $N>0$.
} 
\end{xkn assumption}

\newtheorem{A1 and A2 remark}[order of products remark]{Remark}
\begin{A1 and A2 remark} \label{A1 and A2 rmk}
\textup{
(A1) implies that $\{\Xk\}_k$ has limit points. Observe that the set $\Seto_{\Xo}$ defined in (A2) is of measure zero. Condition (A2) is obviously much weaker than the statement: $\Xknbpij-\Xknij\to 0$ as $n\to\infty$, which is a necessary condition for algorithm convergence and is proved in Proposition \ref{diff iter thm}. (A2) will hold if we have that $\Xkpij-\Xkij\to 0$ as $k\to\infty$, which is much easier to check in practice, but is an overly strong assumption.  
}
\end{A1 and A2 remark}

We have the following convergence theorem:
\newtheorem{final convergence theorem}[L0 vs L1 solutions theorem]{Theorem}
\begin{final convergence theorem} \label{final convergence thm}
\textup{If (A1) and (A2) hold then $\Xk\to\Xd$ as $k\to\infty$, where $\Xd$ is a local minimizer of $\PLL(\cdot)$. 
}
\end{final convergence theorem}

\textit{Proof.} See Appendix B.

\vspace{2mm}

The proof of Theorem \ref{final convergence thm} requires several propositions and lemmas given in Appendix B. We note some of those propositions here and provide a short summary of how they are used. In Proposition \ref{diff iter thm} we show that the difference of the successive iterates converges to zero, a necessary convergence condition. Then, in Proposition \ref{limit points are in F thm} we show that the limit points of the algorithm sequence are fixed points. Ostrowski's result from \cite{Ostrowski73} with Propositions \ref{diff iter thm} and \ref{limit points are in F thm} can subsequently be used to establish that the algorithm sequence converges to a closed and connected subset of fixed points, which is Proposition \ref{general convergence thm}. By Theorem \ref{F is local min thm}, the set of fixed points is a discrete set of local minimizers, and hence the connected subset to which the algorithm sequence converges must be comprised of a single point only, establishing Theorem \ref{final convergence thm}.   

\section{Simulations} \label{simulations section}

Here the performance of the $l_0$ and $l_1$ penalized estimators $\hOm$ of the true precision matrix $\Om_{p\times p}$ are compared. For the $l_0$ penalized estimator we use the proposed CD algorithm, while the $l_1$ penalized estimator is obtained using the $l_q$COV algorithm from \cite{MH14,GMThesis13} with $q=1$, which converges to a unique solution by convexity of the $l_1$-PLL objective function \cite{HSDR11}. Both algorithms are initialized at the same point, as indicated in Section \ref{alg init}. If $\Xo$ denotes the current iterate and $\Xo^+$ denotes the update of $\Xo$ after a single sweep, then these algorithms are terminated when: 
\begin{align}
|\PLL(\Xo)-\PLL(\Xo^+)|/|\PLL(\Xo)|<10^{-8}. \nonumber
\end{align}

\subsection{The Considered Configurations of $\Om$}

\begin{figure}[!ht]
\centering
\begin{minipage}{0.9\linewidth} 
  \centering
  \includegraphics[width=1\linewidth]{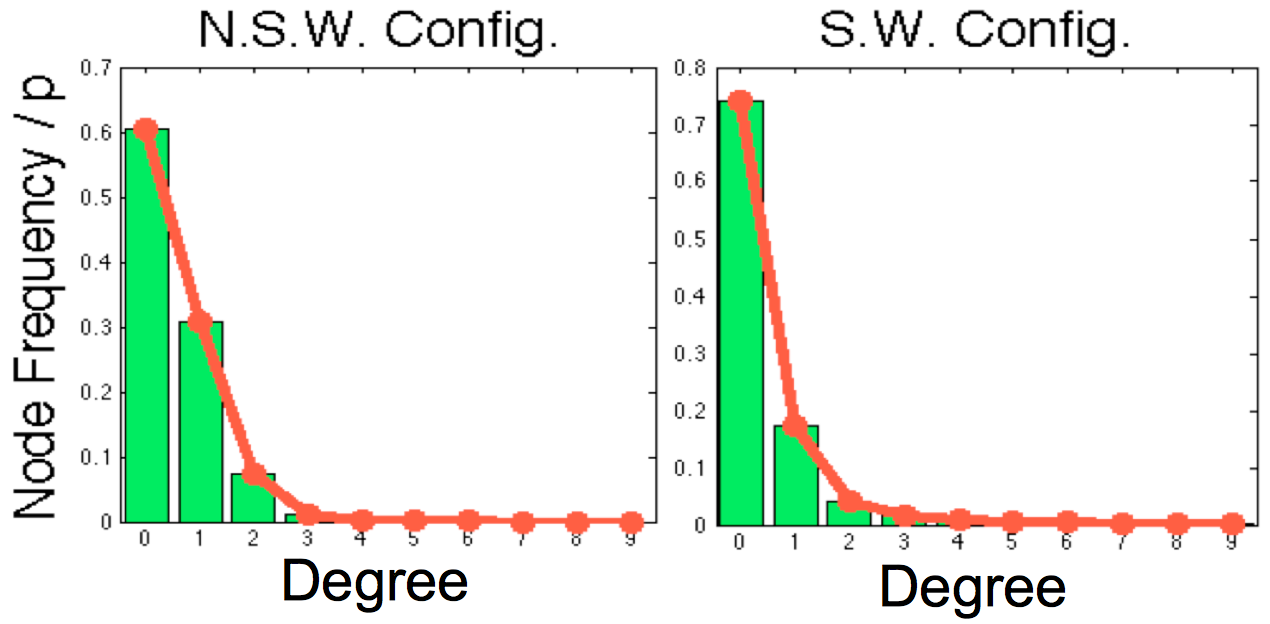}
\end{minipage}
\caption{Histograms showing the degree of node connectivity in the random graph associated with the inverse covariance matrix $\Om_{p\times p}$ for a non-small-world (n.s.w.) and a small-world (s.w.) configuration, respectively. The histograms were generated using $5000$ instances of a sparse $\Om$ containing $50$ off-diagonal non-zeros.}
\label{Node Dist fig}
\end{figure}

We let $p=100$, and consider reconstructing \textbf{s}mall-\textbf{w}orld (\textbf{s.w.}) and \textbf{n}on \textbf{s}mall-\textbf{w}orld (\textbf{n.s.w.}) sparse inverse covariances $\Om$. Non-small-world $\Om$'s are constructed using the Matlab function \texttt{sprandsym}, see \cite{sprandsymFun}. Small-world $\Om$'s are based on the model in \cite{BA99}, and the Matlab code used for construction is from \cite{Power Law}. In these constructions the locations of the zeros and non-zeros in $\Om$ are specified by the adjacency matrix of a sparse random graph. Both n.s.w. and s.w. $\Om$'s have normally distributed off-diagonal non-zeros but the vertex degree distributions of the associated random graphs are very different, see Figure \ref{Node Dist fig}.

\subsection{Varying the Sparsity in $\Om$}

The true sparse inverse covariances $\Om=\Om(\alpha)$ are varied as a function of the sparsity level $\alpha\in[0,1]$, where $\Om(1)$ is the most sparse and $\Om(0)$ is the least sparse matrix. Specifically, we generate $\Om(1)$ and $\Om(0)$ with $\|\Om(1)\|_0=0.015\times p^2$ and $\|\Om(0)\|_0=0.22\times p^2$ using n.s.w. and s.w. models. To generate $\Om(\alpha)$ for any $\alpha\in(0,1)$, we stochastically combine $\Om(1)$ and $\Om(0)$ as follows: Let $r_{ij}$ be independent Bernoulli random variables with the probability parameters:
\begin{align}
p_{ij}=\alpha\cdot\Ind(\Omij(1)\neq 0) + (1-\alpha)\cdot\Ind(\Omij(0)\neq 0) \nonumber
\end{align}
for $i,j=1,\dots,p$. When $r_{ij}\neq 0$, we let:
\begin{align}
\Omij(\alpha)=\alpha\cdot \Omij(1)+(1-\alpha)\cdot \Omij(0). \nonumber
\end{align}

\subsection{The Simulation Procedure} \label{simulation procedure}

\begin{itemize}
\item[$\bullet$] Select a sparsity level  $\alpha\in[0,1]$, and generate $\Om=\Om(\alpha)$. 
\item[$\bullet$] Let $\Lambda$ be the set of 200 linearly equally spaced points between $\lam_{min}$ to $\lam_{max}$. For each $\lam\in\Lambda$ repeat steps (1) to (3) $M=50$ times:
\end{itemize}
\begin{itemize}
\item[(1)] Generate a data set of $n$ i.i.d. multivariate Gaussian random vectors with mean $\mb{0}$ and covariance $\Sig=\Om^{-1}$. 
\item[(2)] Using (1) calculate $\Samp$.
\item[(3)] Compute $\hOm(\lam)$ with the appropriate algorithm, and calculate the Kullback-Leibler (KL) divergence:
\begin{align}
\mukl(\hOm(\lam),\Om)&=-\log\det(\Sig\hOm(\lam))+\tr(\Sig\hOm(\lam))-p \nonumber
\end{align}
\end{itemize}
Note that $\lam_{min}$ and $\lam_{max}$ are functions of the sparsity level $\alpha$, and were empirically chosen such that the global minimizers of $\mukl(\hOm(\lam),\Om)$ (w.r.t. $\lam$) are in the interval $[\lam_{min},\lam_{max}]$.
\begin{itemize}
\item[$\bullet$] Compute the ensemble average oracle performance:
\begin{equation}
\hmukl=\frac{1}{M}\sum_{i=1}^M\min_{\lam}\mukl(\hOm(\lam),\Om) \label{mu hat}
\end{equation}
\end{itemize} 

When $\lambda$ is tuned to give minimum KL divergence, the solution to $\min_{\lam}\mukl$ will be referred to as the penalized ML oracle estimator. The average of these penalized ML oracle estimators (over $M$ trials) will be referred to as the average penalized ML oracle estimator, denoted by $\hOmav$. Lastly, $\hOmav$ and $\hmukl$ superscripted by $l_0$ and $l_1$ correspond to these quantities by minimizing the $l_0$ and $l_1$ penalized PLL objective function, respectively.

In practice we do not have access to $\Om$ so to demonstrate that the proposed method is practically useful we also show results for which $\lam$ has been selected using the computationally efficient \tb{E}xtended \tb{B}ayesian \tb{I}nformation \tb{C}riterion (\tb{EBIC}) \cite{MS314,FD10,CC08}. Unlike the classical methods such as the BIC and Cross Validation, the EBIC is known to work well for sparse graphs when $n$ and $p$ are of similar size \cite{FD10}. To measure the average practical performance we compute:
\begin{align}
\hspace{-2.3mm}\hmukl_{\textup{E}}=\frac{1}{M}\sum_{i=1}^M\textup{KL}(\hOm(\widehat{\lam}),\Om),\ \widehat{\lambda}=\arg\min_{\lam}\textup{EBIC}(\hOm(\lam)) \label{mu hat EBIC}
\end{align} 
where $\textup{EBIC}(\hOm(\lam))$ is stated in \cite{MS314}.

\subsection{Results for Non Small-World (n.s.w.) $\Om$}

\begin{figure}[!ht]
\centering
\begin{minipage}{1.0\linewidth} 
  \centering
  \includegraphics[width=1\linewidth]{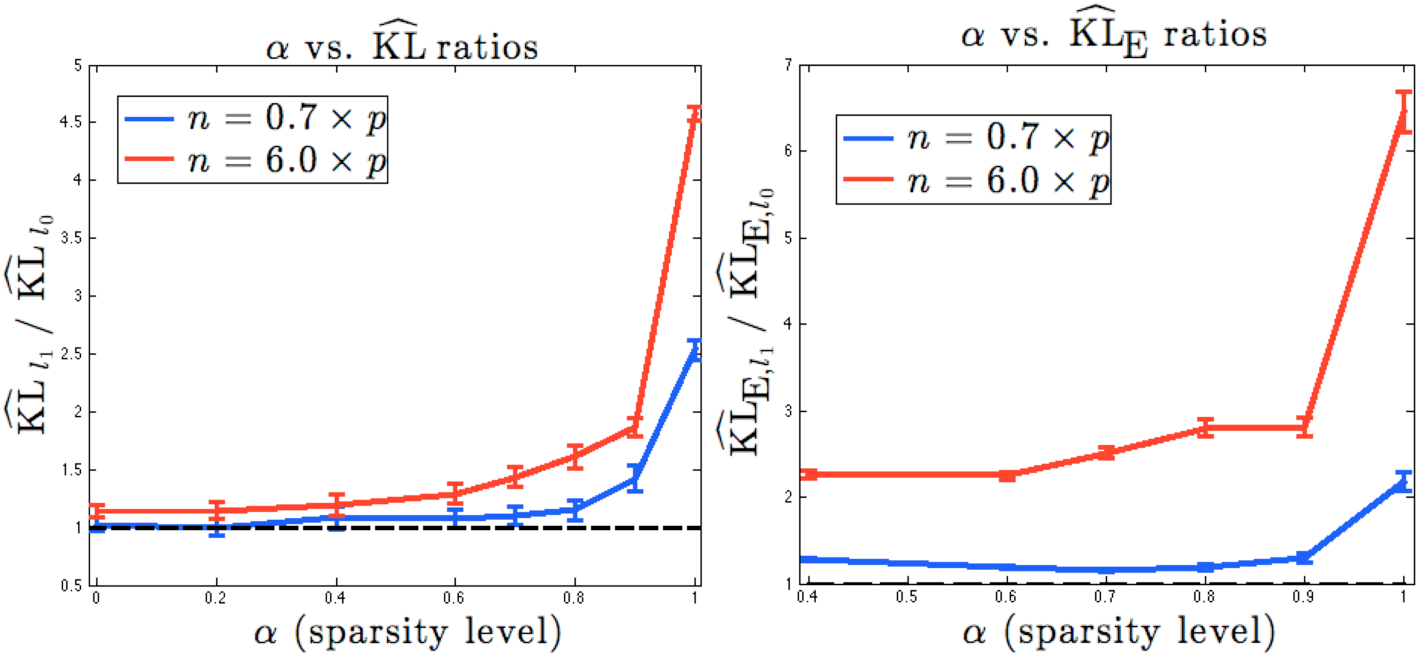}
\end{minipage}
\caption{(Left) Comparing the ratio of the average oracle $\hmukl_{l_0}$ and $\hmukl_{l_1}$ vs. sparsity level $\alpha$ for the non-small-world (n.s.w.) penalized ML estimators. (Right) Comparing the ratio of the average practical $\hmukl_{\textup{E},l_0}$ and $\hmukl_{\textup{E},l_1}$ vs. sparsity level $\alpha$ for which the $l_0$ oracle estimator (left) outperforms the $l_1$ estimator. In both figures the vertical bars have height $1.96\times\textup{SE}$, where SE is the standard error. Note, the $l_0$ advantage improves in the over-determined case where $n=6\times p$.}
\label{Ratio Rand fig}
\end{figure}
Figure \ref{Ratio Rand fig} shows that as the sparsity in $\Om$ increases, the $l_0$ penalized ML estimator outperforms the $l_1$ penalized ML estimator (the error bars are $95\%$ confidence intervals). The performance advantage holds both for under-determined $n=0.7\times p$ and over-determined $n=6\times p$ scenarios.

\begin{figure}[!ht]
\centering
\begin{minipage}{1\linewidth}
  \centering
  \includegraphics[width=1\linewidth]{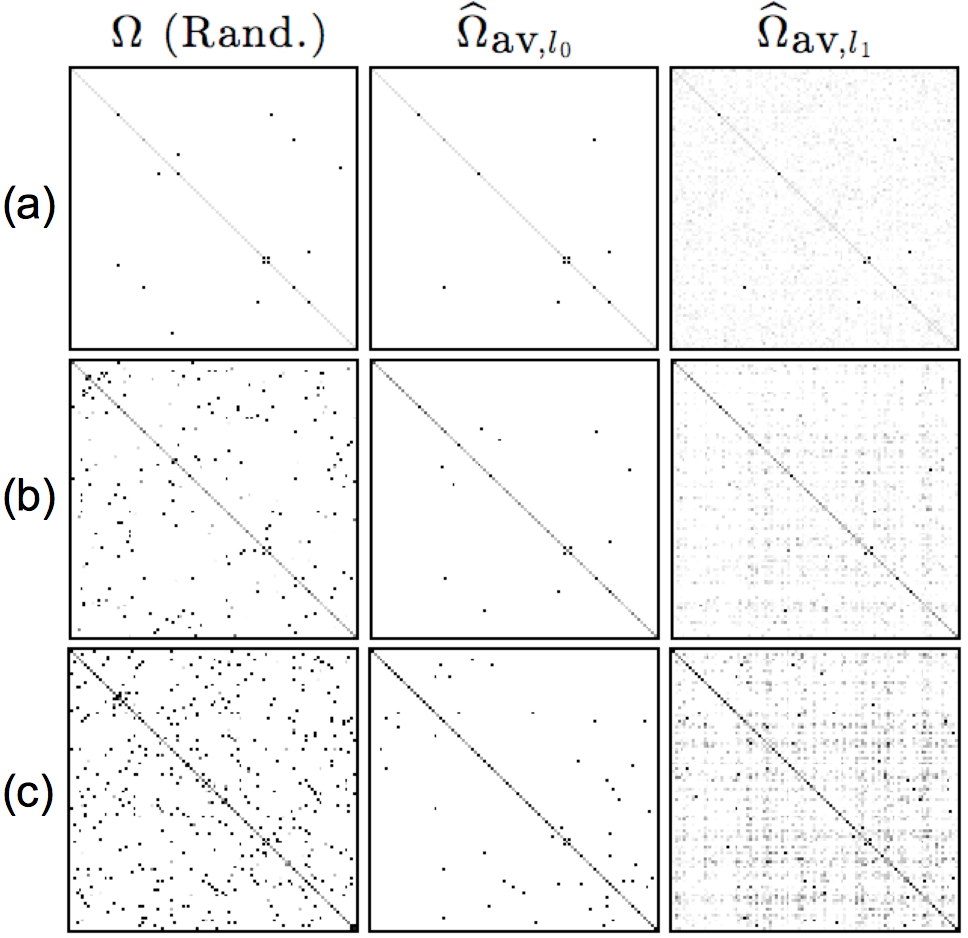}
\end{minipage}%
\caption{The true non-small-world (n.s.w.) $\Om=\Om(\alpha)$ and the corresponding average penalized ML oracle estimators $\hOmav(\alpha)$ with $l_0$ and $l_1$ sparsity penalties when $n=0.7\times p$. The true inverse covariance is a single realization of the n.s.w. configuration. (a) $\alpha=1$, (b) $\alpha=0.9$, and (c) $\alpha=0.8$.  For ease of visualization the inverse covariance estimators have their off-diagonal values magnified 300 times. Notice that the estimates $\hOmavL{1}$ contain many spurious small valued non-zeros unlike the proposed $l_0$ penalized ML estimates $\hOmavL{0}$.}
\label{Rand Omega n=70 fig}
\end{figure}
\begin{figure}[!ht]
\centering
\begin{minipage}{1\linewidth}
  \centering
  \includegraphics[width=1\linewidth]{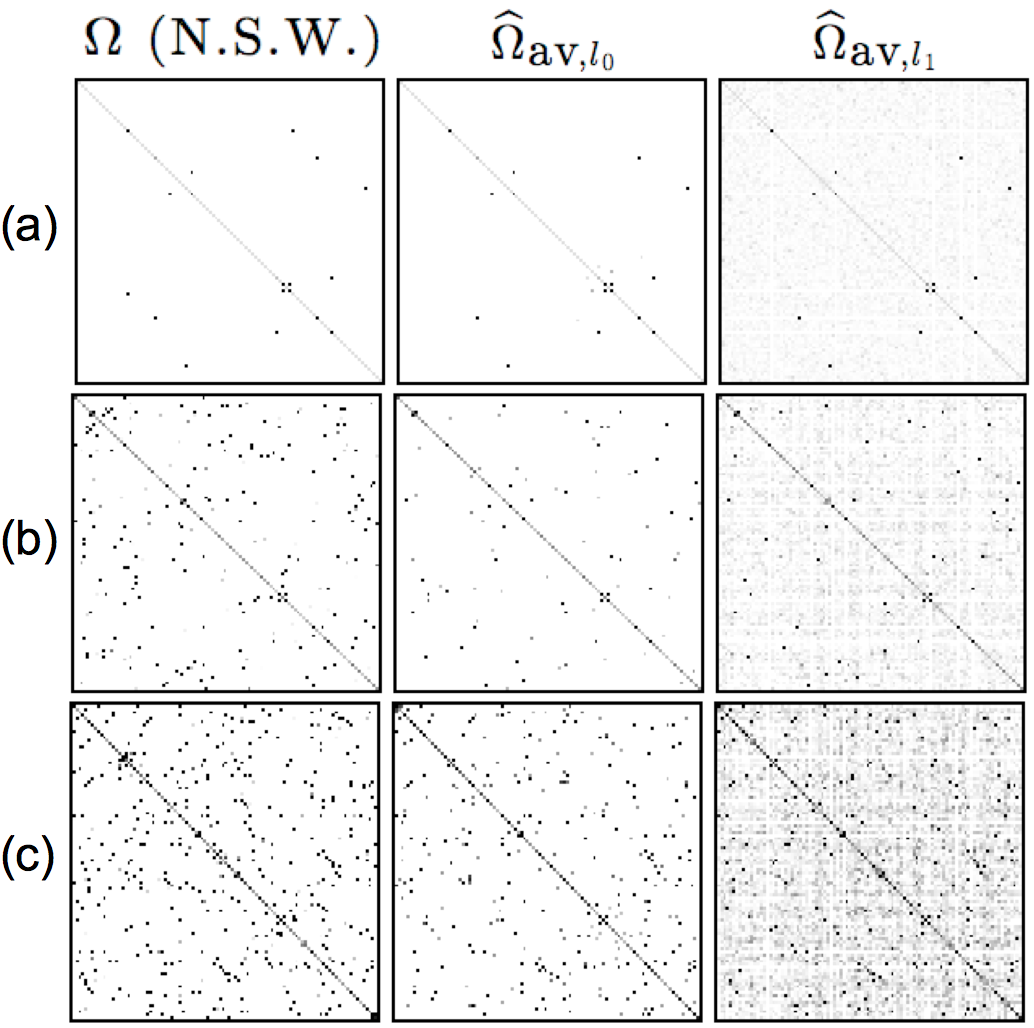}
\end{minipage}%
\caption{The true non-small-world (n.s.w.) $\Om=\Om(\alpha)$ and the corresponding average penalized ML oracle estimators $\hOmav(\alpha)$ with $l_0$ and $l_1$ sparsity penalties when $n=6\times p$. The true inverse covariance is a single realization of the n.s.w. configuration. (a) $\alpha=1$, (b) $\alpha=0.9$, and (c) $\alpha=0.8$. For ease of visualization the inverse covariance estimators have their off-diagonal values magnified 300 times. Notice that the estimates $\hOmavL{1}$ contain many spurious small valued non-zeros unlike the proposed $l_0$ penalized ML estimates $\hOmavL{0}$.}
\label{Rand Omega n=600 fig}
\end{figure}

Figures \ref{Rand Omega n=70 fig} and \ref{Rand Omega n=600 fig} illustrate that the $l_1$ penalized ML oracle estimator has over-estimated the number of non-zero components, and that the $l_0$ penalized ML oracle estimator produces relatively sparser solutions.

\begin{figure}[!ht]
\centering
\begin{minipage}{.9\linewidth}
  \centering
  \includegraphics[width=1\linewidth]{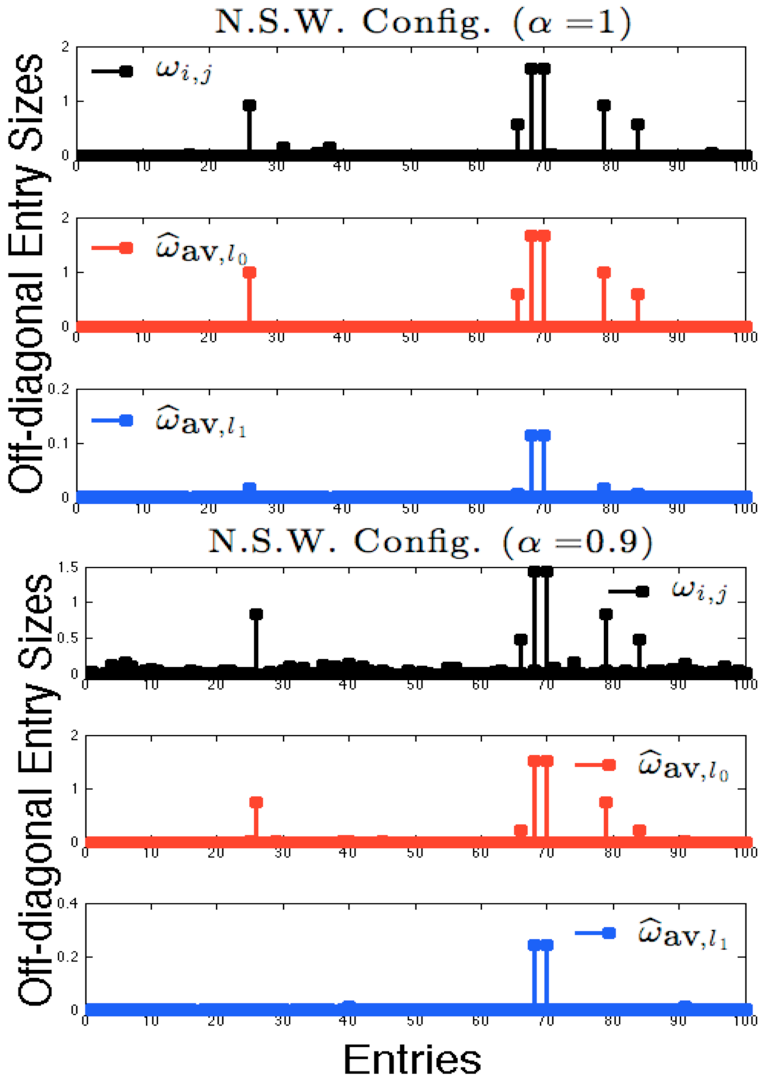} 
\end{minipage}%
\caption{Comparison of amplitudes of the off-diagonal entries in non-small-world (n.s.w.) $\Om=\Om(\alpha)$, $\hOmavL{0}$ and $\hOmavL{1}$, where $n=0.7\times p$. As it can be seen, the $l_1$ exhibits significant shrinkage bias.}
\label{RandEstVals0p7I fig}
\end{figure} 
\begin{figure}[!ht]
\centering
\begin{minipage}{.9\linewidth}
  \centering
  \includegraphics[width=1\linewidth]{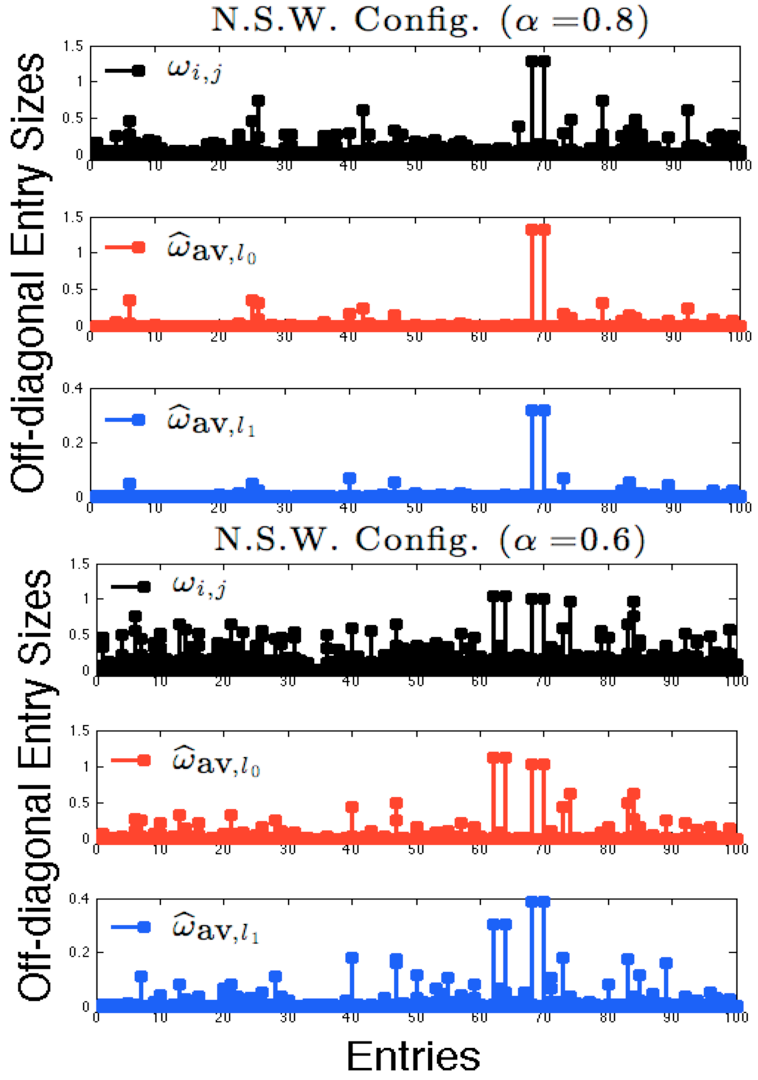} 
\end{minipage}%
\caption{Comparison of amplitudes of the off-diagonal entries in non-small-world (n.s.w.) $\Om=\Om(\alpha)$, $\hOmavL{0}$ and $\hOmavL{1}$, where $n=0.7\times p$. As it can be seen, the $l_1$ exhibits significant shrinkage bias.}
\label{RandEstVals0p7II fig}
\end{figure} 
\begin{figure}[!ht]
\centering
\begin{minipage}{.9\linewidth}
  \centering
  \includegraphics[width=1\linewidth]{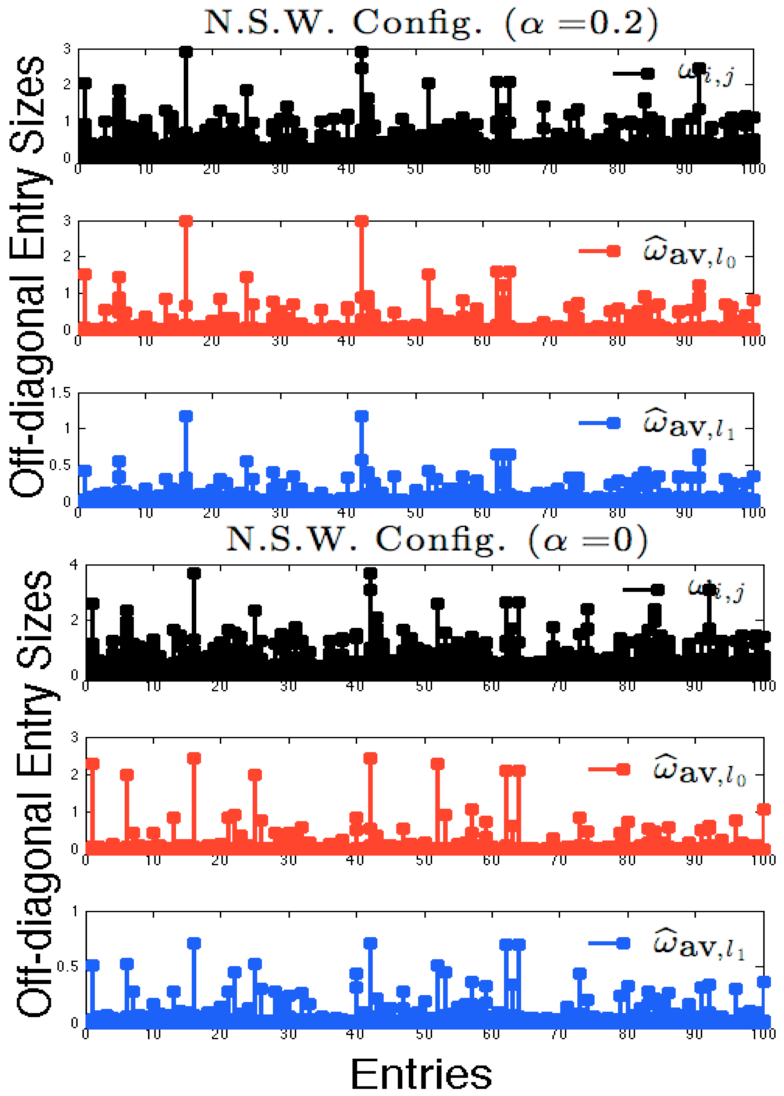} 
\end{minipage}%
\caption{Comparison of amplitudes of the off-diagonal entries in non-small-world (n.s.w.) $\Om=\Om(\alpha)$, $\hOmavL{0}$ and $\hOmavL{1}$, where $n=0.7\times p$. As it can be seen, the $l_1$ exhibits significant shrinkage bias.}
\label{RandEstVals0p7III fig}
\end{figure}

For $n=0.7\times p$, Figures \ref{RandEstVals0p7I fig}, \ref{RandEstVals0p7II fig} and \ref{RandEstVals0p7I fig} confirm the significant shrinkage biases in the larger components of the $l_1$ penalized ML oracle estimator due to the effect of linear penalization in the $l_1$ penalty. We see that no such biases are present in the $l_0$ penalized ML oracle estimator.

\begin{figure}[!ht]
\centering
\begin{minipage}{0.9\linewidth}
  \centering
  \includegraphics[width=1\linewidth]{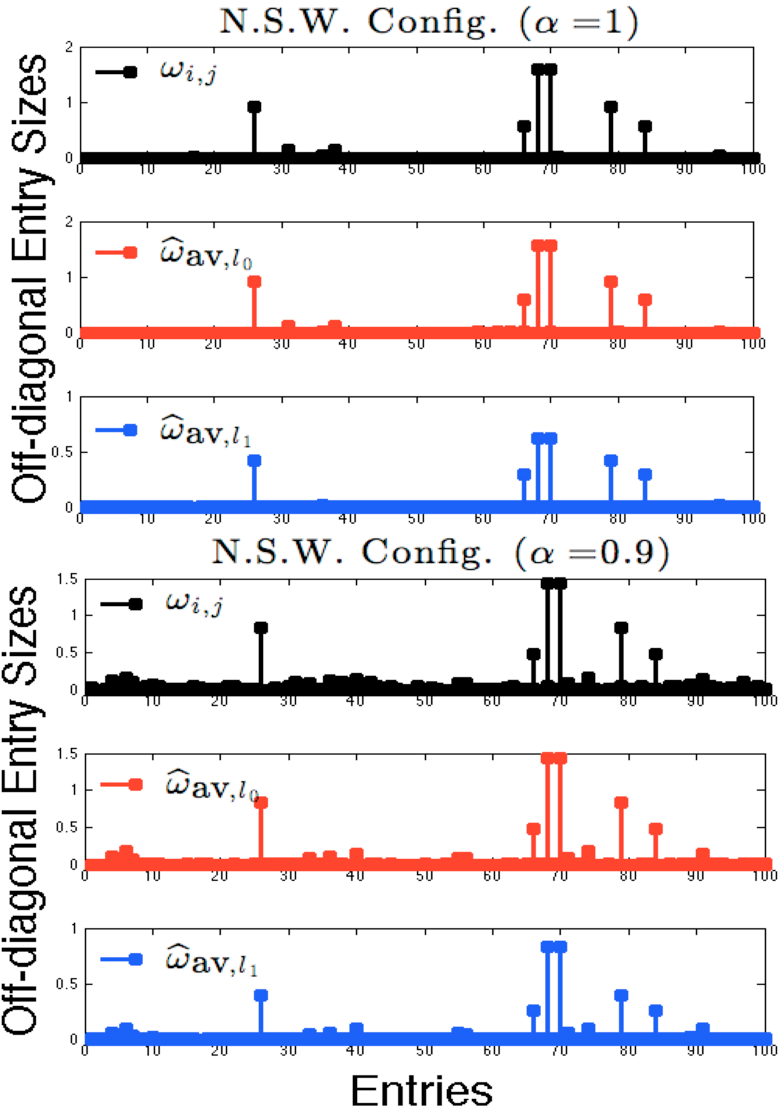} 
\end{minipage}%
\caption{Comparison of amplitudes of the off-diagonal entries in non-small-world (n.s.w.) $\Om=\Om(\alpha)$, $\hOmavL{0}$ and $\hOmavL{1}$, where $n=6\times p$. As it can be seen, the $l_1$ exhibits significant shrinkage bias.}
\label{RandEstVals6p0I fig}
\end{figure} 
\begin{figure}[!ht]
\centering
\begin{minipage}{0.9\linewidth}
  \centering
  \includegraphics[width=1\linewidth]{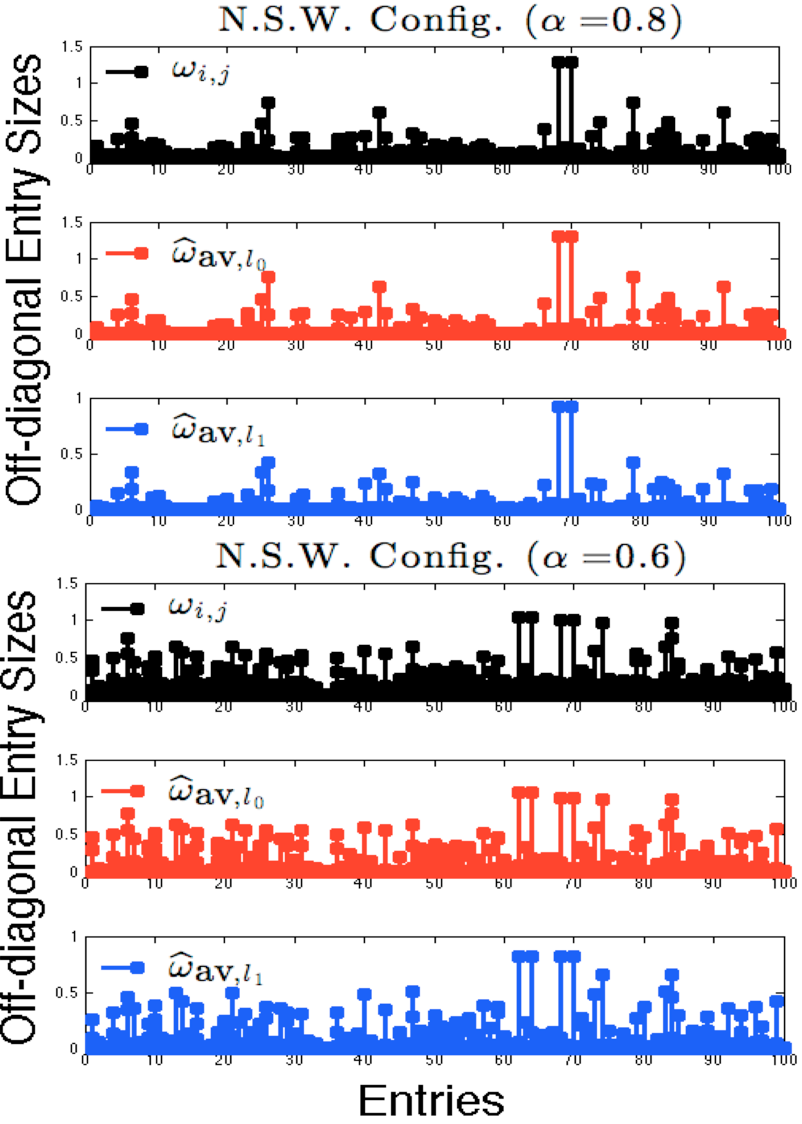} 
\end{minipage}%
\caption{Comparison of amplitudes of the off-diagonal entries in non-small-world (n.s.w.) $\Om=\Om(\alpha)$, $\hOmavL{0}$ and $\hOmavL{1}$, where $n=6\times p$. As it can be seen, the $l_1$ exhibits significant shrinkage bias.}
\label{RandEstVals6p0II fig}
\end{figure} 
\begin{figure}[!ht]
\centering
\begin{minipage}{0.9\linewidth}
  \centering
  \includegraphics[width=1\linewidth]{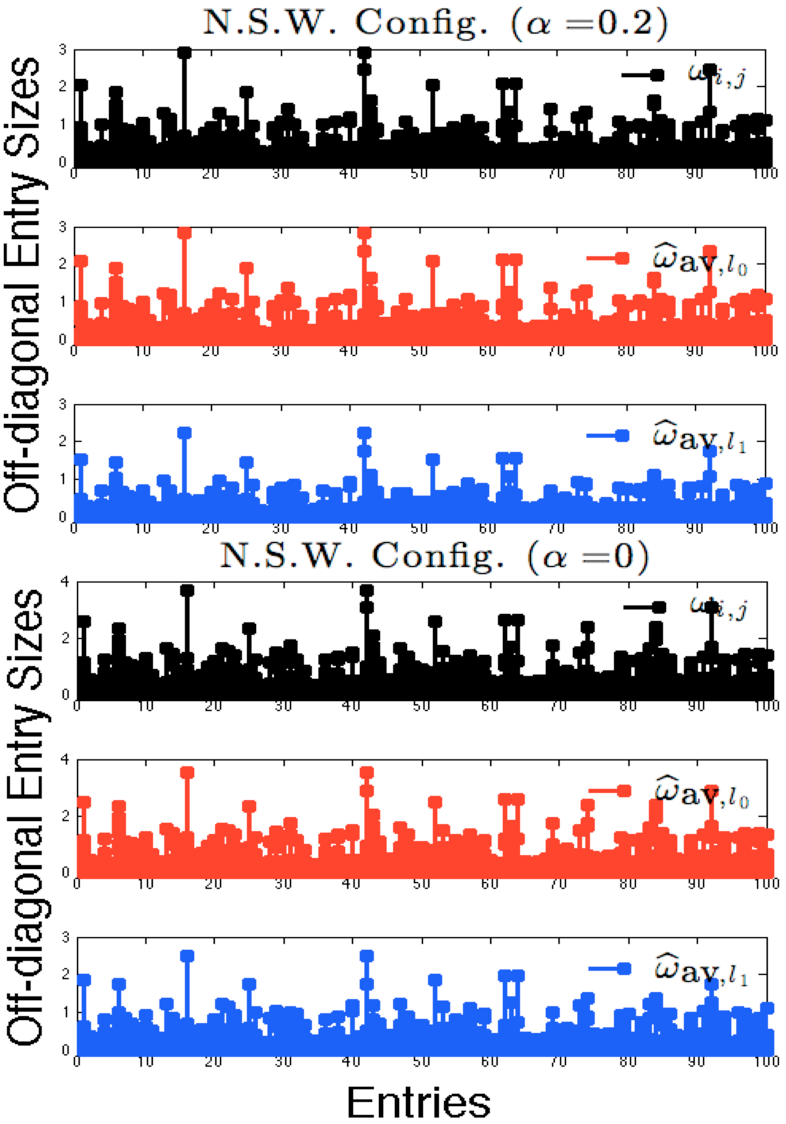} 
\end{minipage}%
\caption{Comparison of amplitudes of the off-diagonal entries in non-small-world (n.s.w.) $\Om=\Om(\alpha)$, $\hOmavL{0}$ and $\hOmavL{1}$, where $n=6\times p$. As it can be seen, the $l_1$ exhibits significant shrinkage bias.}
\label{RandEstVals6p0III fig}
\end{figure} 

For $n=6\times p$, Figures \ref{RandEstVals6p0I fig}, \ref{RandEstVals6p0II fig} and \ref{RandEstVals6p0III fig} also confirm the biases of the $l_1$ penalty. 

Lastly, for $\alpha=1$ and $n=0.7\times p$ we computed averages of ensemble goodness of fit according to KL divergence (\ref{mu hat}).
\begin{figure}[!ht]
\centering
\begin{minipage}{1.0\linewidth}
  \centering
  \includegraphics[width=1\linewidth]{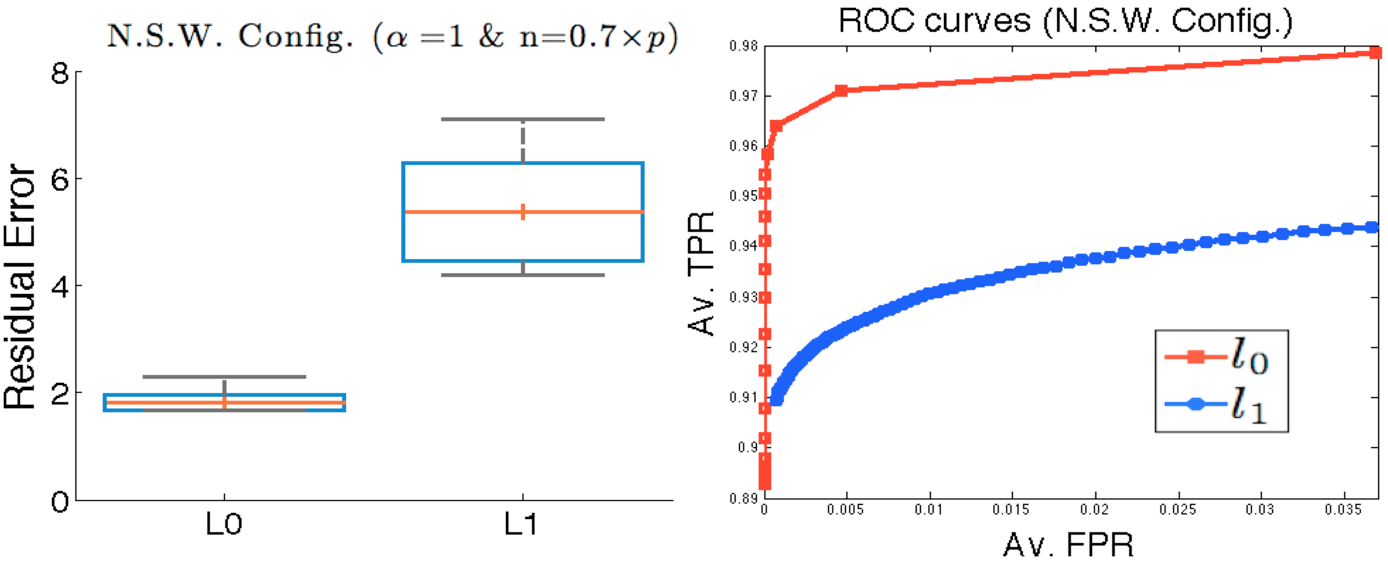} 
\end{minipage}%
\vspace{-0mm}
\caption{(Left) Box and Whisker plot of $\hmukl_{l_0}$ and $\hmukl_{l_1}$ for 15 independent draws from the non-small-world (n.s.w.) ground truth model $\Om=\Om(1)$. The mean $\hmukl_{l_0}$ and mean $\hmukl_{l_1}$ are denoted by the red horizontal line and are given by $1.81$ and $5.38$, respectively. The box represents the standard deviation of $\hmukl$, while the whiskers denote the lowest and highest $\hmukl$ value. (Right) ROC curves plotted using the average true and false positive rates (TPR and FPR). Each TPR and FPR instance is obtained by calculating the TPR and FPR for each $\hOm$ (note that there are $M$ of these $\hOm$) and then taking the average.}
\label{ResErrROCRand fig}
\end{figure} 
Figure \ref{ResErrROCRand fig} shows the results. On the left the ratio between the average $\hmukl_{l_0}$ and $\hmukl_{l_1}$ is $2.97$, which is close to that in Figure \ref{Ratio Rand fig}. On the right, the ROC curves quantitatively establish the superior performance of the $l_0$ penalty. 

\subsection{Results for Small-World (s.w.) $\Om$}

\begin{figure}[!ht]
\centering
\begin{minipage}{1.0\linewidth}
  \centering
  \includegraphics[width=1\linewidth]{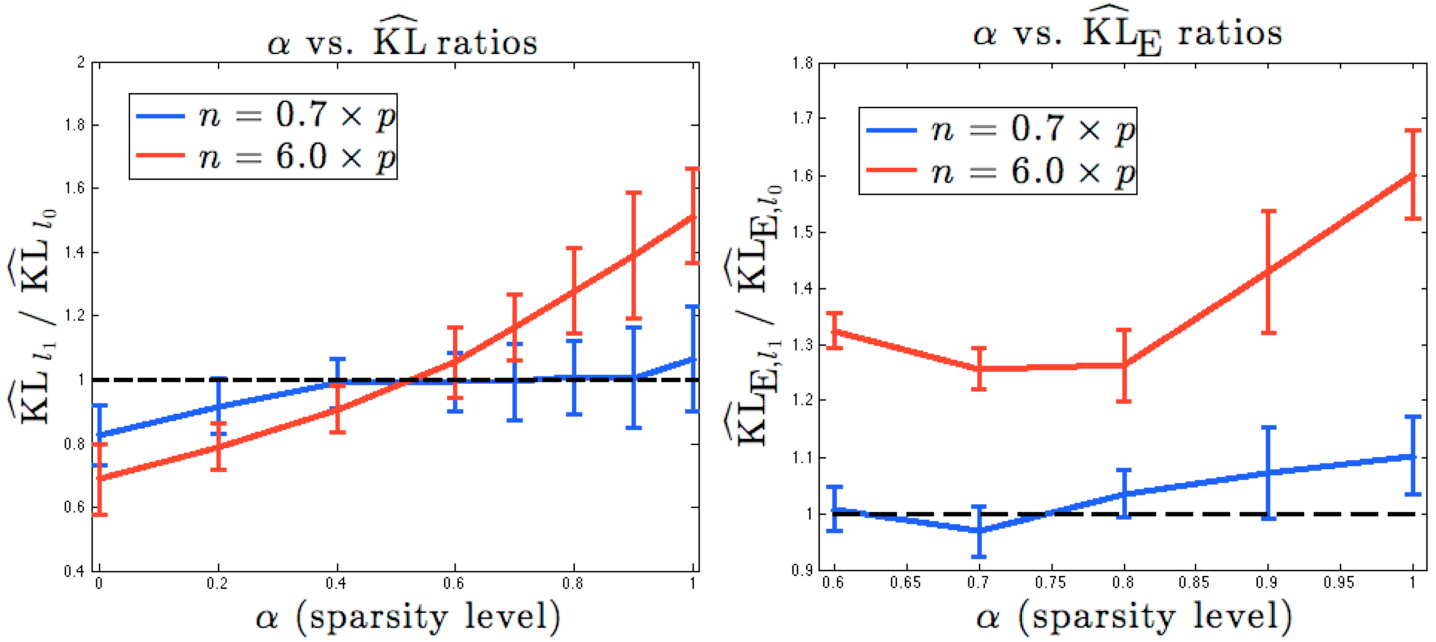}
\end{minipage}%
\caption{(Left) Comparing the ratio of the average oracle $\hmukl_{l_0}$ and $\hmukl_{l_1}$ vs. sparsity level $\alpha\in[0,1]$ for the small-world (s.w.) penalized ML estimators. (Right) Comparing the ratio of the average practical $\hmukl_{\textup{E},l_0}$ and $\hmukl_{\textup{E},l_1}$ vs. sparsity level $\alpha$ for which the $l_0$ oracle estimator (left) outperform the $l_1$ estimator. In both figures the vertical bars have height $1.96\times\textup{SE}$, where SE is the standard error.}
\label{Ratio Star fig}
\end{figure} 

Figure \ref{Ratio Star fig} demonstrates that for a very sparse $\Om=\Om(\alpha)$ the $l_0$ penalized ML estimator has better performance than the $l_1$ penalized ML estimator. This is especially true for $n=6\times p$ case. However, for less sparse scenarios, i.e., for $\alpha<0.5$, we see that the opposite is true, and using the $l_1$ penalty seems to be a better choice in terms of oracle fit in KL divergence. This could be because the proposed $l_0$ approach might be more prone to converge to a local minimizer for lower sparsity levels.

\begin{figure}[!ht]
\centering
\begin{minipage}{1\linewidth}
  \centering
  \includegraphics[width=1\linewidth]{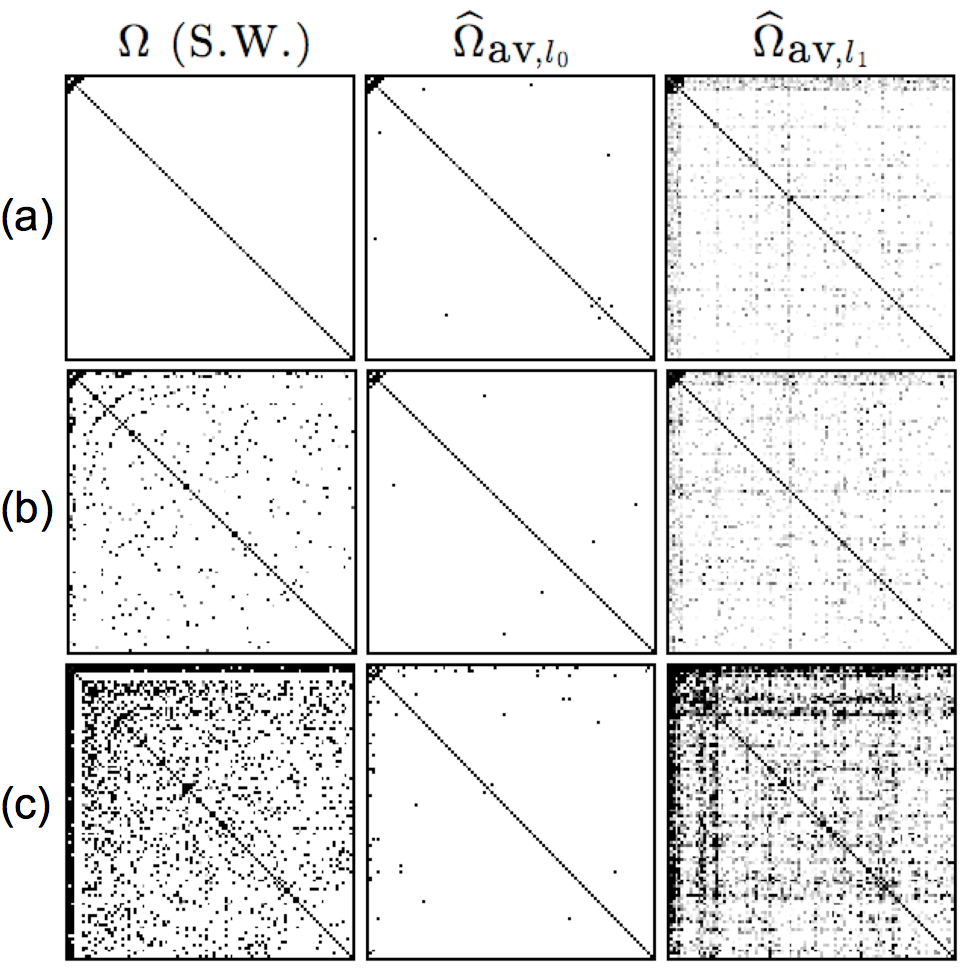}
\end{minipage}%
\caption{The true small-world (s.w.) $\Om(\alpha)$ and the corresponding average penalized ML oracle estimators $\hOmav(\alpha)$ with $l_0$ and $l_1$ sparsity penalties when $n=0.7\times p$. The true inverse covariance is a single realization of the s.w. configuration. (a) $\alpha=1$, (b) $\alpha=0.8$, and (c) $\alpha=0$. The estimators have their off-diagonal values magnified 300 times. As in Figure \ref{Rand Omega n=70 fig} $\hOmavL{1}$ contain many spurious small valued non-zeros unlike the proposed $l_0$ penalized ML estimators $\hOmavL{0}$.}
\label{SW Omega n=70 fig}
\end{figure}
\begin{figure}[!ht]
\centering
\begin{minipage}{1\linewidth}
  \centering
  \includegraphics[width=1\linewidth]{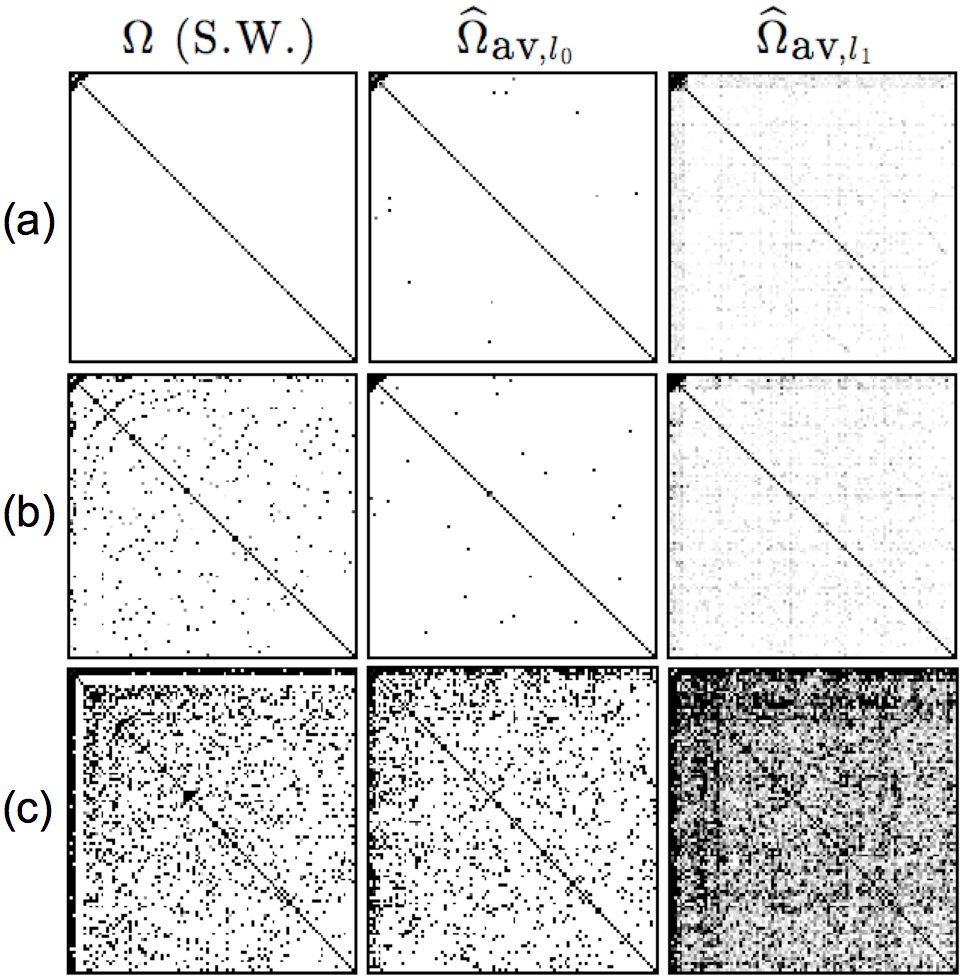}
\end{minipage}%
\caption{The true small-world (s.w.) $\Om(\alpha)$ and the corresponding average penalized ML oracle estimators $\hOmav(\alpha)$ with $l_0$ and $l_1$ sparsity penalties when $n=6\times p$. The true inverse covariance is a single realization of the s.w. configuration. (a) $\alpha=1$, (b) $\alpha=0.8$, and (c) $\alpha=0$. The estimators have their off-diagonal values magnified 300 times. As in Figure \ref{Rand Omega n=600 fig}, $\hOmavL{1}$ contain many spurious small valued non-zeros unlike the proposed $l_0$ penalized ML estimates $\hOmavL{0}$.}
\label{SW Omega n=600 fig}
\end{figure}

Figures \ref{SW Omega n=70 fig} and \ref{SW Omega n=600 fig} show a similar trend as Figure \ref{Rand Omega n=600 fig}, i.e., the average $l_0$  penalized ML oracle estimator is sparser than the average $l_1$ penalized ML oracle estimator, where the latter again contains many more small valued non-zero values.
\begin{figure}[!ht]
\centering
\begin{minipage}{0.9\linewidth}
  \centering
  \includegraphics[width=1\linewidth]{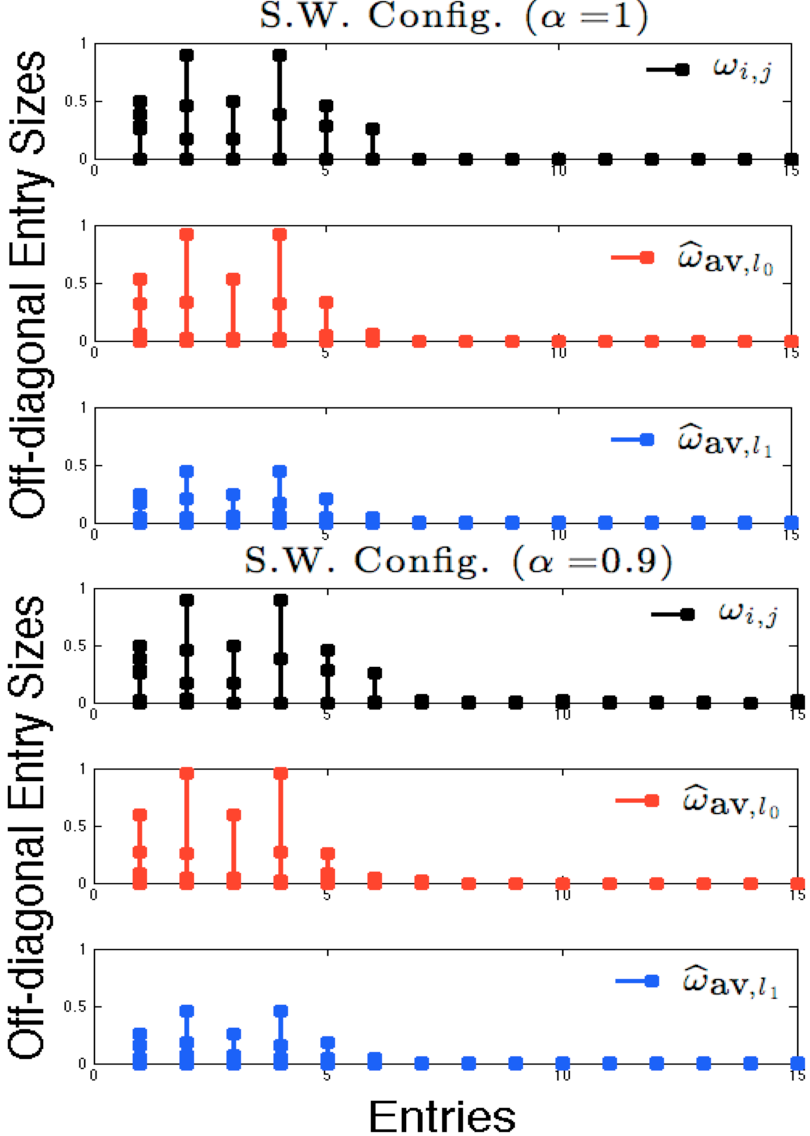}
\end{minipage}%
\vspace{-0mm}
\caption{Comparison of amplitudes of off-diagonal entries in small-world (s.w.) $\Om=\Om(\alpha)$, $\hOmavL{0}$ and $\hOmavL{1}$, where $n=0.7\times p$. The biases of the $l_1$ penalty can easily be seen.}
\label{SWEstVals n=0p7I fig}
\end{figure} 
\begin{figure}[!ht]
\centering
\begin{minipage}{0.9\linewidth}
  \centering
  \includegraphics[width=1\linewidth]{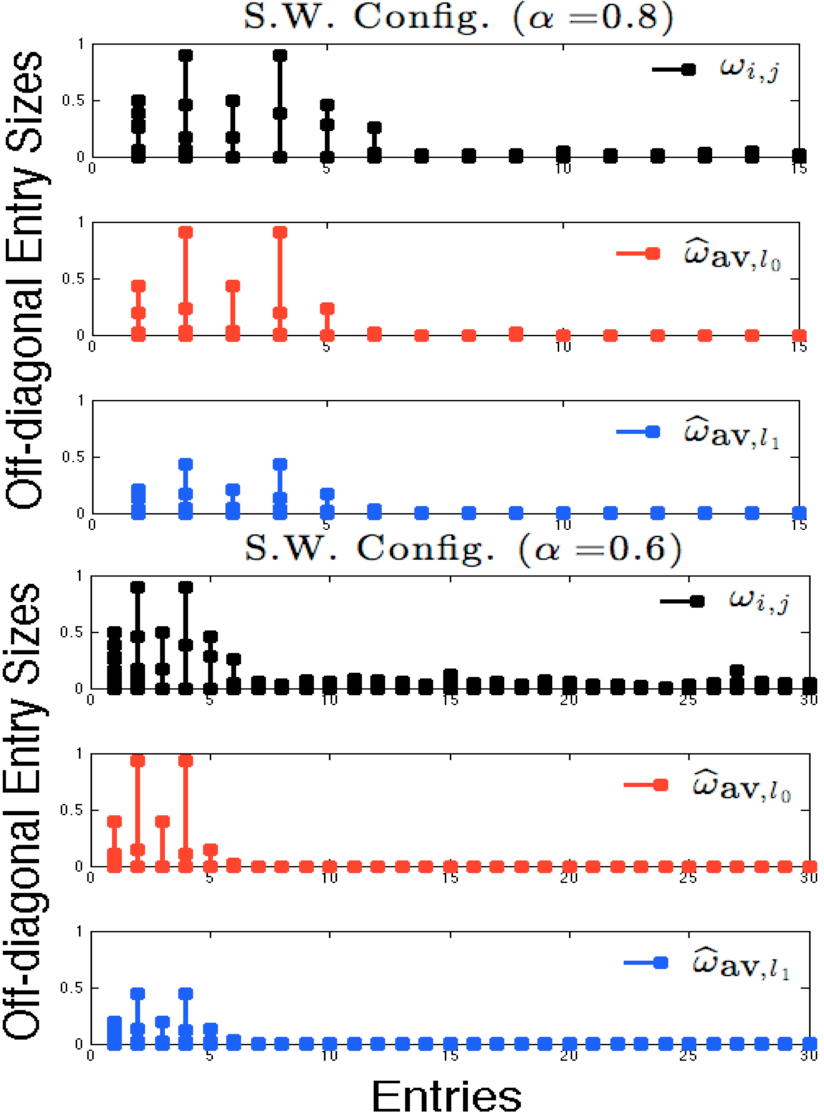}
\end{minipage}%
\vspace{-0mm}
\caption{Comparison of amplitudes of off-diagonal entries in small-world (s.w.) $\Om=\Om(\alpha)$, $\hOmavL{0}$ and $\hOmavL{1}$, where $n=0.7\times p$. The biases of the $l_1$ penalty can easily be seen.}
\label{SWEstVals n=0p7II fig}
\end{figure} 
\begin{figure}[!ht]
\centering
\begin{minipage}{0.9\linewidth}
  \centering
  \includegraphics[width=1\linewidth]{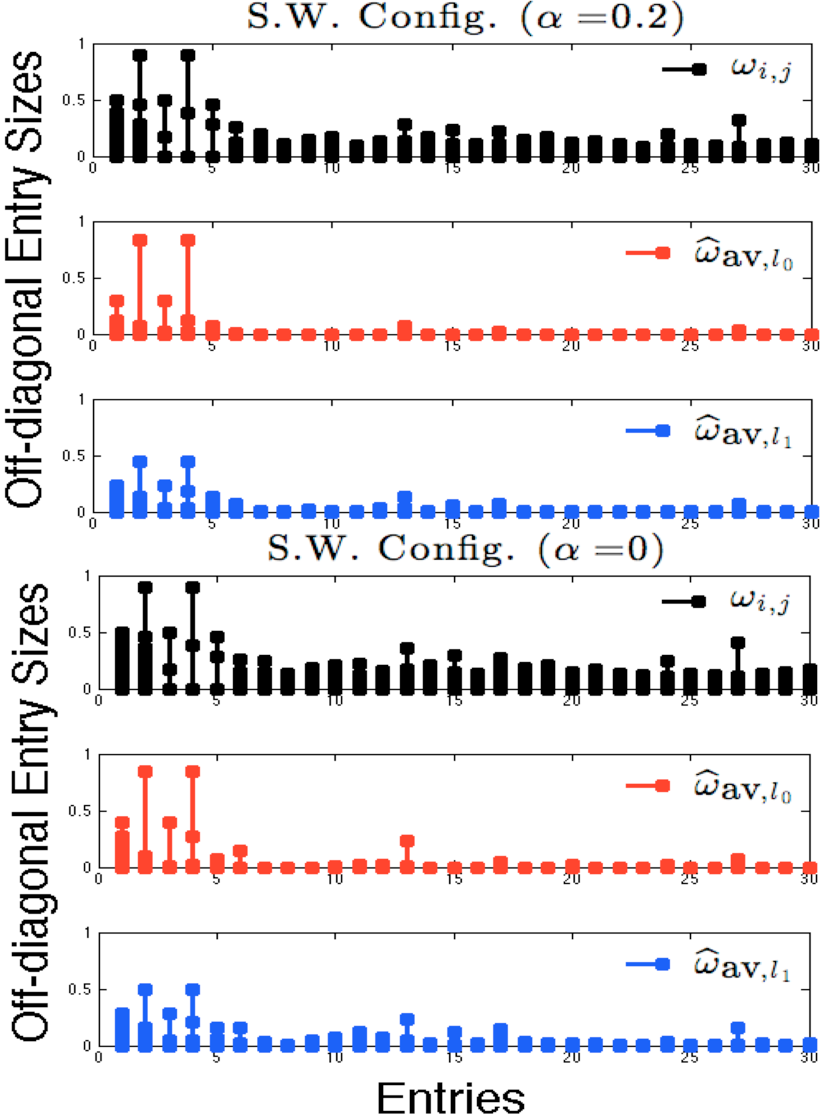}
\end{minipage}%
\vspace{-0mm}
\caption{Comparison of amplitudes of off-diagonal entries in small-world (s.w.) $\Om=\Om(\alpha)$, $\hOmavL{0}$ and $\hOmavL{1}$, where $n=0.7\times p$. The biases of the $l_1$ penalty can easily be seen.}
\label{SWEstVals n=0p7III fig}
\end{figure} 

Figures \ref{SWEstVals n=0p7I fig}, \ref{SWEstVals n=0p7II fig} and \ref{SWEstVals n=0p7III fig} again confirm the biases in the non-zero entries of the average $l_1$ penalized ML oracle estimator unlike the average $l_0$ penalized ML oracle estimator. 

\begin{figure}[!ht]
\centering
\begin{minipage}{0.9\linewidth}
  \centering
  \includegraphics[width=1\linewidth]{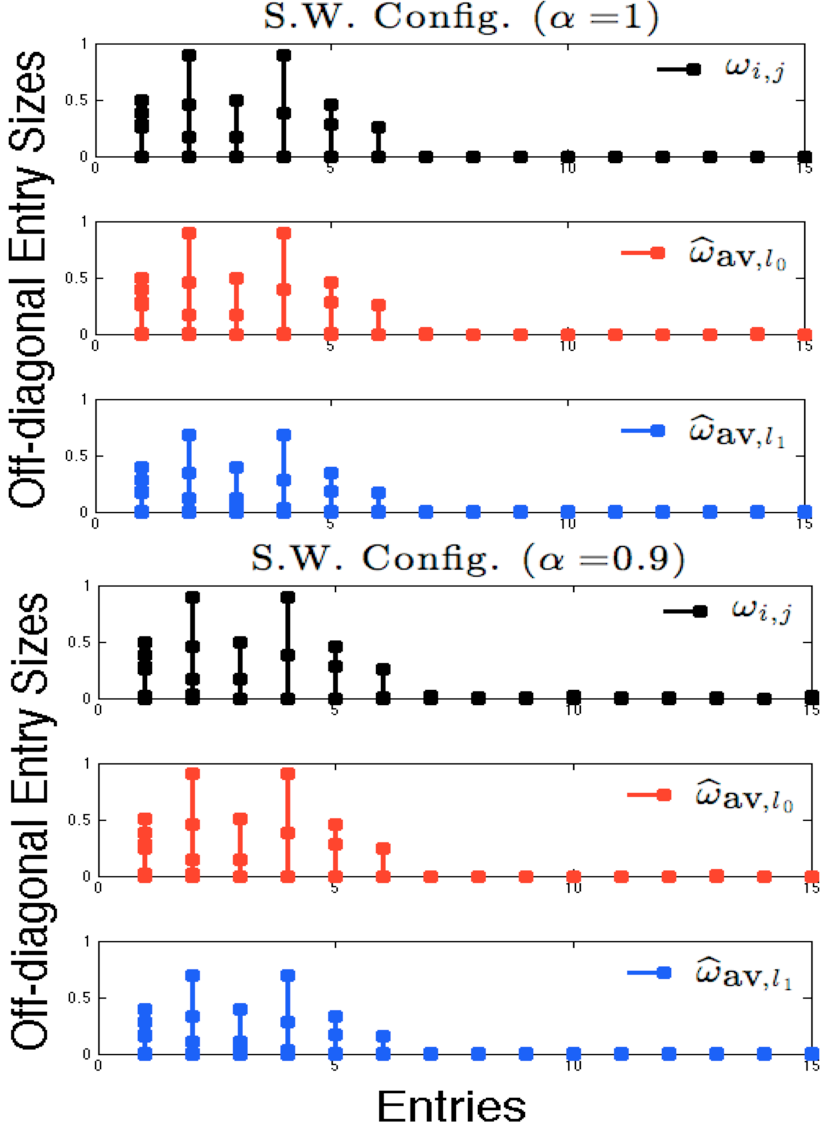}
\end{minipage}%
\vspace{-0mm}
\caption{Comparison of amplitudes of off-diagonal entries in small-world (s.w.) $\Om=\Om(\alpha)$, $\hOmavL{0}$ and $\hOmavL{1}$, where $n=6\times p$. The biases of the $l_1$ penalty can easily be seen.}
\label{SWEstVals n=6p0I fig}
\end{figure} 
\begin{figure}[!ht]
\centering
\begin{minipage}{0.9\linewidth}
  \centering
  \includegraphics[width=1\linewidth]{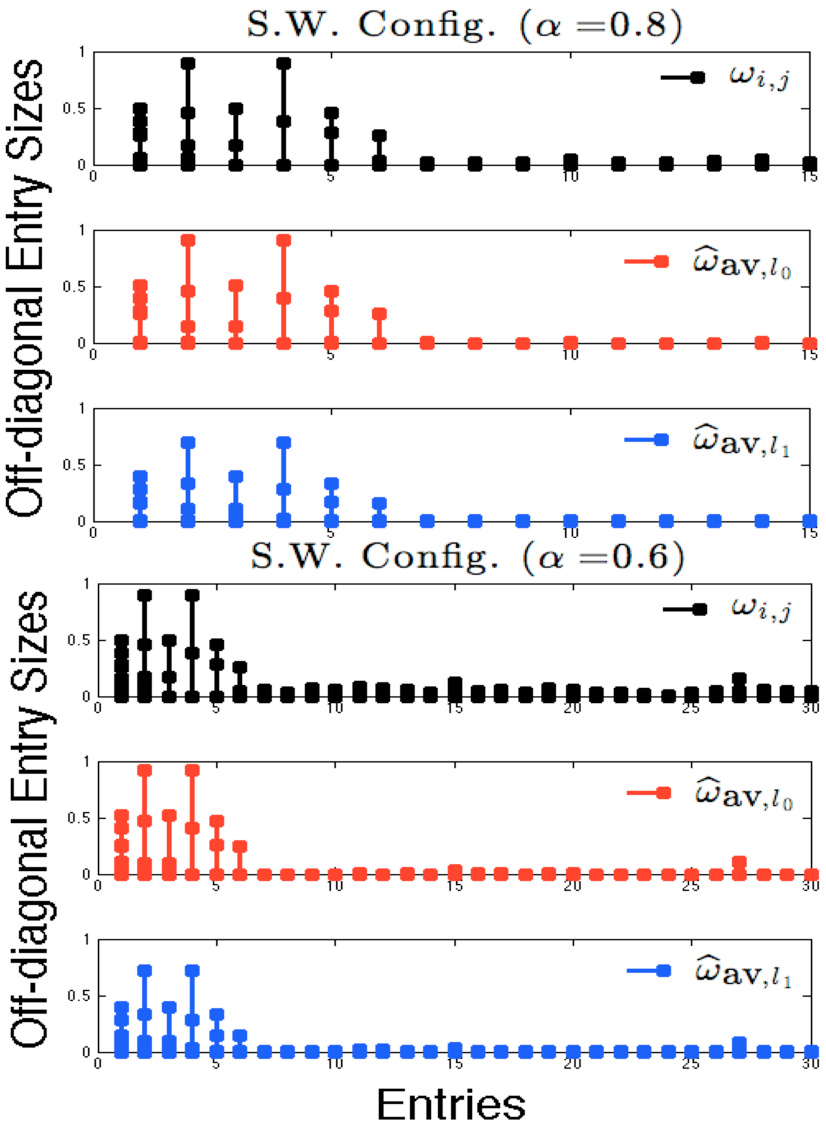}
\end{minipage}%
\vspace{-0mm}
\caption{Comparison of amplitudes of off-diagonal entries in small-world (s.w.) $\Om=\Om(\alpha)$, $\hOmavL{0}$ and $\hOmavL{1}$, where $n=6\times p$. The biases of the $l_1$ penalty can easily be seen.}
\label{SWEstVals n=6p0II fig}
\end{figure} 
\begin{figure}[!ht]
\centering
\begin{minipage}{0.9\linewidth}
  \centering
  \includegraphics[width=1\linewidth]{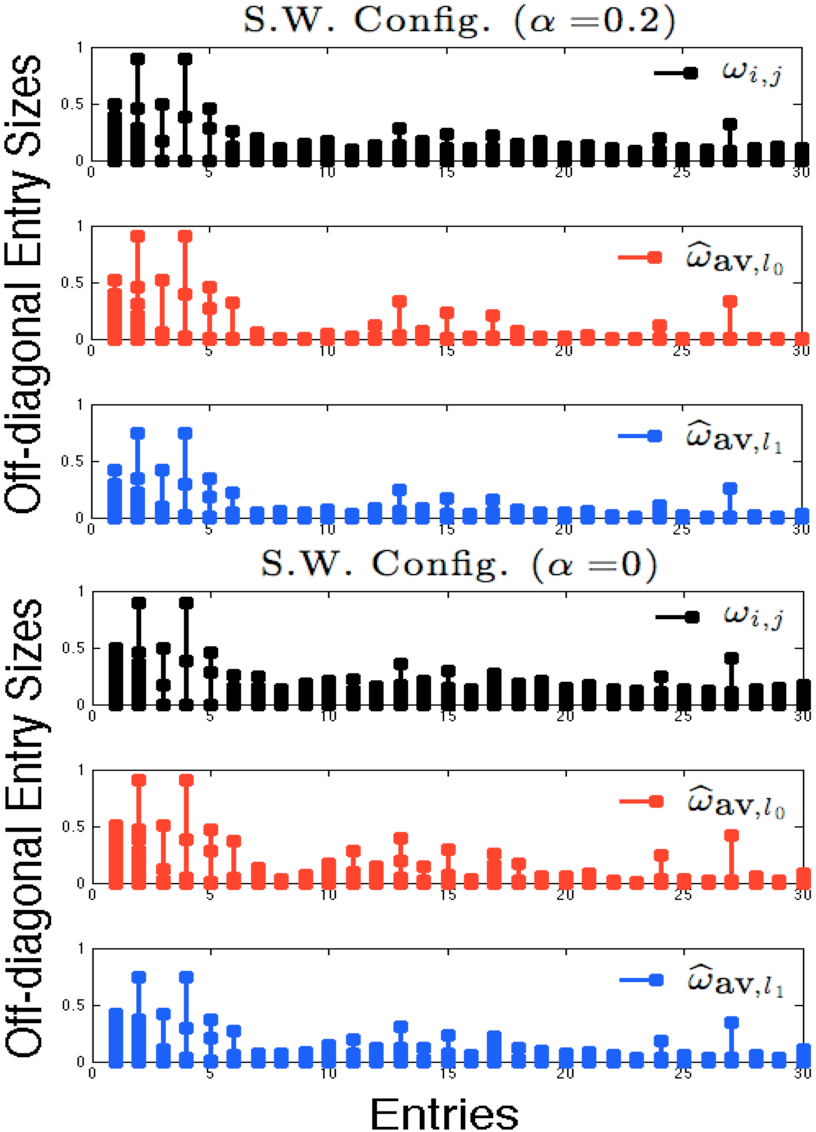}
\end{minipage}%
\vspace{-0mm}
\caption{Comparison of amplitudes of off-diagonal entries in small-world (s.w.) $\Om=\Om(\alpha)$, $\hOmavL{0}$ and $\hOmavL{1}$, where $n=6\times p$. The biases of the $l_1$ penalty can easily be seen.}
\label{SWEstVals n=6p0III fig}
\end{figure} 
For $n=6\times p$, Figures \ref{SWEstVals n=6p0I fig}, \ref{SWEstVals n=6p0II fig} and \ref{SWEstVals n=6p0III fig} indicate that the biases of the $l_1$ penalty are less evident.

Lastly, similarly to the case of n.s.w. $\Om$, we computed the ensemble average performance by repeating the entire simulation procedure in Section \ref{simulation procedure} $15$ times and averaging out the different random draws of s.w. $\Om=\Om(1)$. 
\begin{figure}[!ht]
\centering
\begin{minipage}{1.0\linewidth}
  \centering
  \includegraphics[width=1\linewidth]{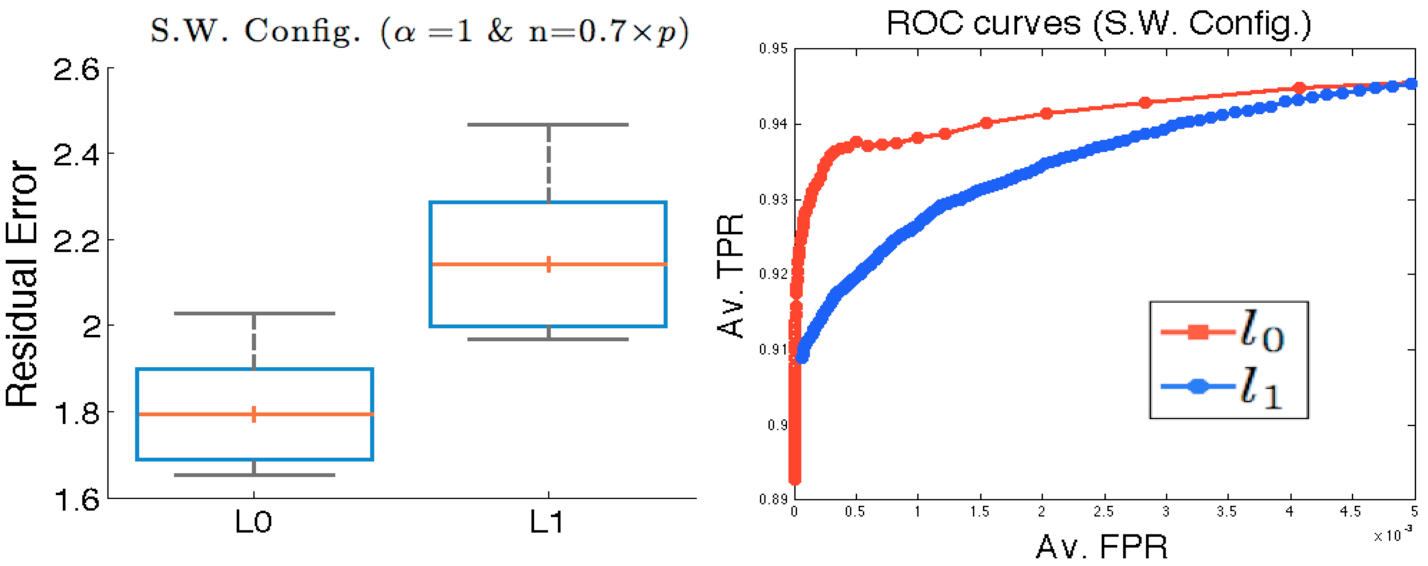} 
\end{minipage}%
\vspace{-0mm}
\caption{(Left) Box and Whisker plot of $\hmukl_{l_0}$ and $\hmukl_{l_1}$ for 15 independent draws from the small-world (s.w.) ground truth $\Om=\Om(1)$. The mean $\hmukl_{l_0}$ and mean $\hmukl_{l_1}$ are denoted by the red horizontal line and are given by $1.79$ and $2.14$, respectively. The box represents the standard deviation of $\hmukl$, while the whiskers denote the lowest and highest $\hmukl$ value. (Right) ROC curves plotted using the average true and false positive rates (TPR and FPR). Each TPR and FPR instance is obtained by calculating the TPR and FPR for each $\hOm$ (note that there are $M$ of these $\hOm$) and then taking the average.}
\label{ResErrROCSW fig}
\end{figure} 
Figure \ref{ResErrROCSW fig} shows the results. On the left the ratio between the average $\hmukl_{l_0}$ and $\hmukl_{l_1}$ is given by $1.20$, which is close to that in Figure \ref{Ratio Star fig}. On the right, the ROC curves show better performance of the $l_0$ penalty.

\section{Conclusion} \label{conclusion section}

We have proposed using the non-convex $l_0$ penalized log-likelihood for estimation of the inverse covariance matrix in Gaussian graphical models as an alternative to the convex $l_1$ penalized log-likelihood approach. We proved that the solutions to the $l_0$ and $l_1$ penalized likelihood maximizations are not generally the same. We developed a novel cyclic descent algorithm for the non-convex optimization and established convergence to a strict local minimizer. 

Comparisons between the penalized Maximum-Likelihood (ML) estimators corresponding to the $l_0$ and the $l_1$ penalty demonstrated two advantages of the proposed $l_0$ penalty for both non small-world and small-world configurations of $\Om$. First, for very sparse inverse covariance we have shown that on average the $l_1$ penalized ML estimators are insufficiently sparse as compared to the $l_0$ penalized ML estimators. Second, we have shown that on average the $l_1$ penalty produces non-zero components that have significantly higher bias due to the shrinkage effect induced by the $l_1$ penalty, which is not induced by the $l_0$  penalty. 

\vspace{1mm}

\textbf{Acknowledgement.} The authors thank Dr. Mila Nikolova for her helpful comments on a late version of this manuscript.

\section*{Appendix A}

For the proofs of results in the paper some standard determinant and matrix inverse identities will be needed. 

In what follows, matrix $\X$ is symmetric and invertible and $\Y=\X^{-1}$. The first result is on the determinant of a perturbed matrix $\X$:
\begin{align}
\det(\X+\del\ei\ei^T)=\det(\X)(1+\del\Yii), \label{det of perturb ii}
\end{align}
where $\ei$ is a unit vector with a $1$ in the $\ith$ entry and $0$ in all other entries. Furthermore:
\begin{align}
\det(\X+\del\Uij\Vij^T)&=\det(\X)\det(\Id+\del\Vij^T\Y\Uij) \nonumber\\[2mm]
&=\det(\X)(-\shurij\del^2+2\Yij\del+1), \label{det of perturb ij}
\end{align} 
where, as defined in (\ref{Uij and Vij def}), $\Uij=[\ei\ \ej]$ and we define:
\begin{equation}
\shurij=\shurij(\Y)=\Yii\Yjj-\Yij^2>0. \label{shur quantity}
\end{equation}
The standard Sherman-Morrison-Woodbury identity gives:
\begin{equation}
(\X+\del\ei\ei^T)^{-1}=\Y-\frac{\del\Y\ei\ei^T\Y}{1+\del\Yii} =\Y-\frac{\del\Yi\Yi^T}{1+\del\Yii} \label{inv of perturb ii}
\end{equation} 
assuming $1+\del\Yii\neq 0$, and:
\begin{align}
&(\X+\del\Uij\Vij^T)^{-1} \nonumber\\[3mm]
&=\Y-\del\Y\Uij(\Id+\del\Vij^T\Y\Uij)^{-1}\Vij^T\Y \nonumber\\[2mm]
&=\Y-\frac{\del\begin{bmatrix} \Yi &\hspace{-2mm} \Yj \end{bmatrix}
\begin{bmatrix} 1+\del\Yij & -\del\Yjj \\ -\del\Yii & 1+\del\Yij \end{bmatrix}
\begin{bmatrix} \Yj^T \\[1mm] \Yi^T \end{bmatrix}}
{-\shurij\del^2+2\Yij\del+1} \label{inv of perturb ij}
\end{align}\\
assuming $-\shurij\del^2+2\Yij\del+1 \neq 0$. 

\section*{Appendix B}

\noindent\tb{Proof of Theorem \ref{L0 vs L1 solutions thm}:} There are two scenarios to consider: (1) the set of local minimizers $\SLo(\lam)$ contains diagonal matrices only, vs. (2) $\SLo(\lam)$ contains at least one matrix with off-diagonal non-zero entries. We cover both simultaneously.

We first derive the necessary optimality condition for a non-zero off-diagonal entry of a local minimizer of (\ref{L0-PLL}): Let $\X\in\SLo(\lam)$ and define $\Y=\X^{-1}$. Denote the set of non-zero entries in $\X$ by:
\begin{align}
\Zc(\X)=\{(i,j):\Xij\neq 0\}, \nonumber
\end{align}
which is non-empty by assumption. Now, since $\X$ is a local minimizer, by definition there exists an $\eps>0$ such that:
\begin{equation}
\PLL(\X+\Del)\geq\PLL(\X) \textup{ for any }\Del \textup{ with }\|\Del\|_F<\eps, \label{local min expression}
\end{equation}
where $\Del=[\delij]$ is a symmetric matrix perturbation. Letting $(i,j)\in\Zc(\X)$ and consider:
\begin{align}
\Del=
\begin{cases}
\delii\ei\ei^T & \textup{ if }i=j \nonumber\\[2mm]
\delij\Uij\Vij^T & \textup{ otherwise. } \nonumber
\end{cases}
\end{align}
By substituting $\del=\delij$ into (\ref{det of perturb ij}) and (\ref{shur quantity}), we have:

\vspace{2mm}

\noindent (i) If $i=j$, then:
\small
\begin{align}
\PLL(\X+\Del)-\PLL(\X)&=-\log\det(\X+\Del) + \tr(\Samp(\X+\Del)) \nonumber\\[2mm]
& +\lam\|\X+\Del\|_0 + \log\det(\X) - \tr(\Samp\X)  \nonumber\\[2mm]
&-\lam\|\X\|_0 \nonumber\\[2mm]
&=-\log\left\{\det(\X)(1+\delii\Yii)\right\}  + \tr(\Samp\Del)  \nonumber\\[2mm]
&+ \log\det(\X) + 2\lam\underbrace{\Id(\Xii+\del\neq 0)}_{=1}  \nonumber\\
&- 2\lam\underbrace{\Id(\Xii\neq 0)}_{=1} =-\log(1+\delii\Yii)+\Sampii\delii. \nonumber
\end{align} \normalsize
(ii) If $i\neq j$, then:
\small
\begin{align}
\PLL(\X+\Del)-\PLL(\X)&=-\log\det(\X+\Del) + \tr(\Samp(\X+\Del)) \nonumber\\[2mm]
&+ \lam\|\X+\Del\|_0 + \log\det(\X) - \tr(\Samp\X) \nonumber\\[2mm]
& - \lam\|\X\|_0  \nonumber\\[2mm]
&=-\log\left\{\det(\X)(-\shurij\delij^2+2\Yij\delij+1)\right\} \nonumber\\[2mm]
&+ \tr(\Samp\Del) +\log\det(\X)   \nonumber\\[2mm]
&+2\lam\Ind(\Xij+\delij\neq 0) -2\lam\underbrace{\Id(\Xij\neq 0)}_{=1} \nonumber\\
&=-\log(-\shurij\delij^2+2\Yij\delij+1) \nonumber\\[2mm] 
&+2\delij\Sampij + 2\lam\Ind(\Xij+\delij\neq 0)-2\lam, \nonumber
\end{align} \normalsize
where $\shurij=\shurij(\Y)$ and $\shurij(\cdot)$ is defined in (\ref{shur quantity}). 

\vspace{2mm}

Suppose that:
\begin{align}
|\delij|<\min\{|\Xij|,\eps/2\}. \nonumber
\end{align}
Since $|\delij|<|\Xij|$, we have:
\begin{align}
\Ind(\Xij+\delij\neq 0)=\Ind(\Xij\neq 0)=1 \textup{ for }i\neq j, \nonumber
\end{align}
and (\ref{local min expression}) is equivalent to:
\small
\begin{align}
0 \leq & f(\delij) \nonumber\\[2mm]
=&
\begin{cases}
-\log(1+\delii\Yii)+\Sampii\delii & \textup{if }i=j \nonumber\\[2mm]
-\log(-\shurij\delij^2+2\Yij\delij+1)+2\delij\Sampij & \textup{otherwise} \nonumber
\end{cases}
\end{align} \normalsize
for any $|\delij|<\min\{|\Xij|,\eps/2\}$. Noting that $f(0)=0$, and $f(\delij)\geq 0$ in a small region around $\delij=0$, we must have $f^\prime(0)=0$. Thus, by differentiating $f(\delij)$ and letting $\delij\to 0$:
\begin{equation}
-\Yij+\Sampij=0, \textup{ for any }(i,j)\in\Zc(\X). \label{non-zero opt cond L0}
\end{equation}
This is the necessary condition for $\X$ to be in $\SLo(\lam)$. To finish the proof we relate  (\ref{non-zero opt cond L0}) to $\hOmLon(\mu)=[\hOmLonij]$. Defining:
\begin{align}
\hSigLon(\mu)=\hOmLon^{-1}(\mu)=[\hSigLonij], \nonumber
\end{align}
it is well known \cite{FHT08} that the necessary and sufficient condition for $\hOmLonij\neq 0$ is:
\begin{align}
\begin{cases}
-\hSigLonij+\Sampij + \mu=0 & \textup{if }i=j \\[2mm]
-\hSigLonij+\Sampij + \mu\sgn(\hOmLonij)=0  & \textup{otherwise}.
\end{cases} \label{non-zero opt cond L1}
\end{align}
For $\hOmLon(\mu)\in\SLo(\lam)$ to be true, (\ref{non-zero opt cond L0}) and (\ref{non-zero opt cond L1}) need to hold simultaneously for some $\mu>0$. But, this is not possible, which completes the proof. \qed

\vspace{2mm}

The following simple lemma will be useful for the subsequent proofs:

\newtheorem{quadratic=0 theorem}{Lemma}
\begin{quadratic=0 theorem} \label{quadratic=0 thm}
\textup{
Suppose $\Xknij\to\Xdij$ as $n\to\infty$. Define:
\begin{align}
q(\Xknij)=-\shurknij(\Xknij)^2-2\Yknij\Xknij+1 \label{q function}
\end{align}
and suppose $q(\Xdij)=0$. Then for a large enough $N>0$ we have:
\begin{align}
\Map(\Xknij)=\Xsknij \textup{ for all }n>N \nonumber
\end{align}
}
\end{quadratic=0 theorem}

\textit{Proof.} $q(\Xknij)$ is continuous w.r.t. $\Xknij$, and $q(\Xdij)=0$ can in general be reached in an oscillating fashion as $n\to\infty$. Namely, we might have $q(\Xknij)\leq 0$ for some $n$, and $q(\Xknij)>0$ for some other $n$. Since $\Map(\Xknij)=\Xsknij$ for those $n$ corresponding to $q(\Xknij)\leq 0$, we now only need to focus on $n$ for which $q(\Xknij)>0$. We proceed by recalling (\ref{pll(0) expression}):
\begin{align}
\pllknij(0)=-\log\det(\Xkn)-\log(q(\Xknij)). \nonumber 
\end{align}
Since $\Xknij\to\Xdij$ we have that: 
\begin{align}
\Xkn\to\Xd, \nonumber
\end{align}
and this implies:
\begin{align}
\log\det(\Xkn)\to\log\det(\Xd), \nonumber
\end{align}
which is finite. However, having $q(\Xknij)\to 0$ implies:
\begin{align}
\log(q(\Xknij))\to\infty. \nonumber
\end{align}
Therefore, $\pllknij(0)\to\infty$ as $n\to\infty$.

\vspace{2mm}

Next, recalling (\ref{pllc(Xsij) expression}) we have that:
\begin{align}
\pllcknij(\Xsknij)&=-\log\det(\Xkn) - \log(\bq(\Xknij)) \nonumber\\[2mm]
&\hspace{3.8mm} + 2\Sampij\Xsknij + 2\lam, \nonumber
\end{align}
where:
\begin{align}
\bq(\Xknij)&=-\shurknij\del(\Xsknij)^2+2\Yknij\del(\Xsknij)+1 \nonumber\\[4mm]
&=
\begin{cases}
1+\frac{(\Yknij)^2}{\shurknij} & \textup{if }\Sampij=0 \\[4mm]
\frac{\sqrt{\left(\shurknij\right)^2+4\Sampij^2\Yknii\Yknjj}-\shurknij}{2\Sampij^2} & \textup{otherwise}. \nonumber
\end{cases}
\end{align} 
The second equality for $\bq(\Xknij)$ can easily be shown using the results in Section \ref{mij when i neq j}. Now, notice that $\bq(\Xknij)>0$ for all $n$ and $\shurknij,\Yknii,\Yknjj$, which are themselves strictly positive for all $n$. Since:
\begin{align}
\shurknij\to\shurdij>0,\ \Yknii\to\Ydii>0, \textup{ and }\Yknjj\to\Ydjj>0 \nonumber
\end{align} 
we must have that:
\begin{align}
\bq(\Xsknij)\to\bq^{\bullet}>0. \nonumber
\end{align}
Also note that $\Xsknij\to\Xsdij$, which is finite by the same reason that $\bq^\bullet$ is finite (see the definition of $\Xsij$). This means:
\begin{align}
\pllcknij(\Xsknij)\to c^\bullet, \nonumber
\end{align}
which is finite. Thus, there has to exist a large enough $N>0$ such that:
\begin{align}
\pllknij(0)>\pllcknij(\Xsknij) \textup{ for all }n>N, \nonumber
\end{align}
implying $\Map(\Xknij)=\Xsknij$ for all $n>N$. \qed

\vspace{3mm}

\noindent\tb{Proof of Theorem \ref{Map is fixed point thm}:} Firstly, note that if $\Xkij\to\Xdij$ then we must have $\Xkpij\to\Xdij$ as well. As a result:
\begin{align}
\Map(\Xkij)\to\Xdij, \nonumber
\end{align}
and so, all that needs to be shown is that $\Map(\Xkij)\to\Map(\Xdij)$.

\vspace{1mm}

When $i=j$, we have $\Map(\Xkij)=\Xskij$, which is continuous w.r.t. $\Xkij$. Thus, 
\begin{align}
\Xkij\to\Xdij \textup{ implies }\Map(\Xkij)\to\Map(\Xdij). \nonumber
\end{align}

\vspace{1mm}

Now suppose $i\neq j$ and consider the fact that:
\begin{align}
q(\Xkij)\to q(\Xdij), \nonumber
\end{align}
where $q(\cdot)$ is defined in (\ref{q function}). There are two cases:

\vspace{4mm}

$\mb{[C_1]:}$ Suppose $q(\Xdij)\leq 0$. By the definition of $\Map(\cdot)$ and Lemma \ref{quadratic=0 thm} (with $k_n=k$), we have $\Map(\Xkij)=\Xskij$ for all $k>K$. Since $\Xskij$ is a continuous function of $\Xkij$ we must have $\Map(\Xkij)\to\Map(\Xdij)$.

\vspace{4mm}

$\mb{[C_2]:}$ Suppose $q(\Xdij)>0$, and noting that $\pllKij(\Xskij)=\pllcKij(\Xskij)$ define:
\begin{align}
\Phi_{\Xk,ij}(\Xkij)=\pllKij(0)-\pllcKij(\Xskij), \label{Phi function}
\end{align} 
which is continuous w.r.t. $\Xkij$, in which case:
\begin{align}
\Phi_{\Xk,ij}(\Xkij)\to\Phi_{\Xd,ij}(\Xdij). \nonumber
\end{align}
There are now two scenarios:
 
\vspace{1mm}
 
\noindent\hspace{5mm} (i) $\Phi_{\Xd,ij}(\Xdij)\neq 0$: This implies that for a sufficiently large $K>0$, we have:
\begin{align}
\Phi_{\Xk,ij}(\Xkij)<0 \textup{ \underline{or} } \Phi_{\Xk,ij}(\Xkij)>0 \textup{ for all }k>K. \nonumber
\end{align}
In the former case, $\Map(\Xkij)=0$, and in the latter case $\Map(\Xkij)=\Xskij$ for all $k>K$. These are both continuous w.r.t. $\Xkij$ implying $\Map(\Xkij)\to\Map(\Xdij)$.

\vspace{2mm}

\noindent\hspace{5mm} (ii) $\Phi_{\Xd,ij}(\Xdij)=0$: Since $\Map(\Xkij)\to\Xdij$, for a large enough $K$ we have to have $\Phi_{\Xk,ij}(\Xkij)$ approach $0$ either from below or from above for all $k>K$. So, suppose:
\begin{align}
\Phi_{\Xk,ij}(\Xkij)<0 \textup{ for all } k>K. \nonumber
\end{align}
Then $\Map(\Xkij)=0$ for all $k>K$, which implies:
\begin{align}
\Map(\Xkij)\to 0, \nonumber
\end{align}
and thus, $\Xdij=0$. So, using the definition of $\Map(\cdot)$, at $\Xdij$ we have:
\begin{equation} 
\Map(\Xdij)=\Xsdij\cdot\Ind(\Xdij\neq 0)=\Xsdij\cdot 0=0, \nonumber
\end{equation}
implying $\Map(\Xkij)\to\Map(\Xdij)$. 

\vspace{2mm}

Alternatively, suppose:
\begin{align}
\Phi_{\Xk,ij}(\Xkij)>0 \textup{ for all } k>K. \nonumber
\end{align}
Then $\Map(\Xkij)=\Xskij$ for all $k>K$, which implies:
\begin{align}
\Map(\Xkij)\to\Xsdij, \nonumber
\end{align}
and thus, $\Xdij=\Xsdij$. Therefore, using the definition of $\Map(\cdot)$, at $\Xdij$ we have:
\begin{equation} 
\Map(\Xdij)=\Xsdij\cdot\Ind(\Xdij\neq 0)=\Xsdij\cdot 1=\Xsdij, \nonumber
\end{equation}
implying $\Map(\Xkij)\to\Map(\Xdij)$. This completes the proof. \qed

\vspace{3mm}

The proof of Theorem \ref{F is local min thm} requires Lemmas \ref{property 1 of F thm} and \ref{property 2 of F thm}:

\vspace{2mm}

\newtheorem{property 1 of F theorem}[quadratic=0 theorem]{Lemma}
\begin{property 1 of F theorem} \label{property 1 of F thm}
\textup{Suppose $\Xo\in\Fix$ and $\Xoij\neq 0$. Define:
\begin{align}
\Yo=\Xo^{-1}. \nonumber
\end{align}
Then, $\Yoij=\Sampij$. 
}
\end{property 1 of F theorem}

\textit{Proof.} Having $\Xo\in\Fix$ implies $\Xoij=\Map(\Xoij)$. 

\vspace{2mm}

When $i=j$ we have $\Map(\Xoii)=\Xsii$, where $\Xsij$ is from (\ref{mii value}). This implies:
\begin{align}
\Xoii=\Xoii+\frac{\Yoii-\Sampii}{\Yoii\Sampii}, \nonumber
\end{align}
which reduces to $\Yoii=\Sampii$ after simplification.

\vspace{2mm}

When $i\neq j$, having $\Xoij\neq 0$ implies $\Map(\Xoij)=\Xsij$, where $\Xsij$ is defined in (\ref{mij value 1}) and (\ref{mij value 2}). As a result, we have the following equation:
\begin{align}
\Xoij&=\Xoij \nonumber\\[2mm]
&+
\begin{cases}
\frac{\Yoij}{\shuroij} & \textup{if }\Sampij=0 \nonumber\\[3mm]
\frac{\Yoij}{\shuroij}+\frac{\shuroij-\sqrt{\shuroij^2+4\Sampij^2\Yoii\Yojj}}{2\shuroij\Sampij} & \textup{otherwise}, \nonumber
\end{cases}
\end{align}
where $\shuroij=\shurij(\Yo)$ and $\shurij(\cdot)$ is from (\ref{shur quantity}). After simplification we can easily obtain that $\Yoij=\Sampij$. \qed

\vspace{3mm}

\newtheorem{property 2 of F theorem}[quadratic=0 theorem]{Lemma}
\begin{property 2 of F theorem} \label{property 2 of F thm}
\textup{Suppose $\Xo\in\Fix$ and $\Xoij=0$, where $i\neq j$. Letting $\Yo=\Xo^{-1}$, there exists $\del>0$ that depends on $\lam$, $\Xo$ and $\Samp$ such that:
\begin{align}
|\Yoij-\Sampij|\leq\del. \nonumber
\end{align} 
}
\end{property 2 of F theorem}

\textit{Proof.} As in Lemma \ref{property 2 of F thm}, $\Xo\in\Fix$ implies $\Xoij=\Map(\Xoij)$. Having $\Xoij=0$ means $\Map(\cdot)$ is given by (\ref{algorithm map 1}) and:
\begin{equation}
\left.\pllij(0)\right|_{\Xoij=0}\leq\left.\pllij(\Xsij)\right|_{\Xoij=0}. \label{xij=0 cond}
\end{equation}
Recall the following standard inequalities:
\begin{align}
& \log(a)\geq \frac{(a-1)}{a}, \textup{ for any }a>0 \label{log inequality}\\[2mm]
& \sqrt{a^2+b}-a\geq\sqrt{b}-a, \textup{ for any }a,b\geq 0 \label{sqrt inequality}
\end{align}
Dealing with (\ref{xij=0 cond}) requires two cases:

\vspace{2mm}

$\mb{[C_1]:}$ $\Sampij=0$. It can easily be shown that (\ref{xij=0 cond}) reduces to: 
\begin{align}
\log\left(1+\frac{\Yoij^2}{\shuroij}\right)\leq 2\lam. \nonumber
\end{align}
So, using (\ref{log inequality}) with:
\begin{align}
a=1+\frac{\Yoij^2}{\shuroij}, \nonumber
\end{align}
we obtain that:
\begin{align}
|\Yoij|\leq\sqrt{2\lam\Yoii\Yojj}. \nonumber
\end{align}
Since $|\Yoij-\Sampij|=|\Yoij|$ the proof is complete for $\mb{[C_1]}$.

\vspace{2mm}

$\mb{[C_2]:}$ $\Sampij\neq 0$. It can easily be shown that (\ref{xij=0 cond}) reduces to:\\
\begin{equation}
\underbrace{\log\left(\frac{\sqrt{\sqoij-\shuroij}}{2\Sampij^2}\right)}_{\textup{\large($\star$)}}+
\underbrace{\frac{\sqrt{\sqoij}-\shuroij-2\Sampij\Yoij}{\shuroij}}_{\textup{\large($\star\star$)}}\leq 2\lam, \nonumber
\end{equation}
where 
\begin{align}
\sqoij=\shuroij^2+4\Sampij^2\Yoii\Yojj>0, \nonumber
\end{align}
noting that $\shuroij>0$ as well. So, using (\ref{log inequality}) with:
\begin{align}
a=\frac{\sqrt{\sqoij-\shuroij}}{2\Sampij^2}, \nonumber
\end{align}
we have:
\begin{equation}
\textup{($\star$)}\geq\frac{\sqrt{\sqoij}-\shuroij-2\Sampij^2}{\sqrt{\sqoij}-\shuroij}>-\frac{2\Sampij^2}{\sqrt{\sqoij}-\shuroij}. \label{log inequality 2}
\end{equation} 
The last inequality in (\ref{log inequality 2}) comes from the fact that:
\begin{align}
\sqrt{\sqoij}-\shuroij>0. \nonumber
\end{align}
Next, substituting:
\begin{align}
a=\shuroij \textup{ and } b=4\Sampij^2\Yoii\Yojj \nonumber
\end{align}
in  (\ref{sqrt inequality}), we obtain: 
\begin{align}
\sqrt{\sqoij}\geq 2|\Sampij|\sqrt{\Yoii\Yojj}. \nonumber
\end{align}
Thus:
\begin{align}
\textup{($\star\star$)}&\geq\frac{2|\Sampij|\sqrt{\Yoii\Yojj}-\shuroij-2\Sampij\Yoij}{\shuroij} \nonumber\\[2mm]
&=\frac{2|\Sampij|\sqrt{\Yoii\Yojj}-(\Yoii\Yojj-\Yoij^2)-2\Sampij\Yoij}{\shuroij} \nonumber\\[2mm]
&=\frac{(\Yoij-\Sampij)^2}{\shuroij}-\frac{(|\Sampij|-\sqrt{\Yoii\Yojj})^2}{\shuroij}. \label{inequality 3}
\end{align}
As a result, (\ref{log inequality 2}), (\ref{inequality 3}) and the fact that $\textup{($\star$)}+\textup{($\star\star$)}\leq 2\lam$ imply (after re-arrangement) that: $|\Yoij-\Sampij|<\del$ for some $\del>0$. This completes the proof. \qed

\vspace{3mm}

\noindent\tb{Proof of Theorem \ref{F is local min thm}:} Let $\Y=\X^{-1}$ and introduce:
\begin{align}
\LL(\X)=-\log\det(\X)+\tr(\Samp\X). \nonumber
\end{align}
The Hessian is equal to:
\begin{align}
\nabla^2\LL(\X)=\Y\kronp\Y\succ 0. \nonumber
\end{align}
Since any eigenvalue of $\nabla^2\LL(\X)$ is a continuous function of $\X$, there exists a small neighbourhood of $\X$, denoted by:
\begin{align}
\Nh_{\eps_0}(\X)=\{\Xp=\X+\Del:0\leq\|\Del\|_F<\eps_0\}, \nonumber
\end{align}
such that $\nabla^2\LL(\Xp)\succ 0$ for all $\Xp\in\Nh_{\eps_0}(\X)$. In other words, there exists a constant $\mu>0$ such that:
\begin{align}
\nabla^2\LL(\Xp)\succeq\mu\Id \textup{ for all } \Xp\in\Nh_{\eps_0}(\X), \nonumber
\end{align}
which in turn implies that $\LL(\cdot)$ is strongly convex in $\Nh_{\eps_0}(X)$. Recalling the standard inequality for a strongly convex function:
\begin{align}
\LL(\X+\Del)&\geq\LL(\X)+\tr(\nabla\LL(\X)\Del)+\frac{1}{2}\mu\|\Del\|_F^2 \nonumber\\
&=\LL(\X)+\sum_{ij}\frac{1}{2}\mu\delij^2+(-\Yij+\Sampij)\Delij, \label{LL inequality}
\end{align}
where $\|\Del\|_F<\eps_0$. The equality in (\ref{LL inequality}) comes from using:
\begin{align}
\nabla\LL(\X)=-\Y+\Samp. \nonumber
\end{align}
Now, using the fact that $\X\in\Fix$ implies $\Xij\in\Fixij$, we introduce the following sets:
\begin{align}
& \Zs_{\X}=\{(i,j):i\neq j,\ \Xij=0\},\nonumber\\[2mm] 
& \Zcs_{\X}=\{(i,j):(i,j)\notin\Zs_{\X}\} \nonumber
\end{align}
Using (\ref{LL inequality}) we obtain: 
\begin{align}
\PLL(\X+\Del)\geq\PLL(\X)+\Res(\Del), \nonumber
\end{align}
where it can be easily shown that:
\begin{align}
\Res(\Del)&=\sum_{ij}\frac{1}{2}\mu\Delij^2+(-\Yij+\Sampij)\Delij + \nonumber\\
&\hspace{10mm}+\lam\Ind(\Xij+\Delij\neq 0)-\lam\Ind(\Xij\neq 0) \nonumber\\[2mm]
&=\hspace{-2mm}\sum_{(i,j)\in\Zs}\hspace{-1mm}\frac{1}{2}\mu\Delij^2+(-\Yij+\Sampij)\Delij + \lam\Ind(\Delij\neq 0) \nonumber\\[2mm]
&+\hspace{-2mm}\sum_{(i,j)\in\Zcs}\hspace{-1mm}\frac{1}{2}\mu\Delij^2+(-\Yij+\Sampij)\Delij+ \nonumber\\
&\hspace{10mm}+\lam\Ind(\Xij+\Delij\neq 0)-\lam. \nonumber
\end{align}
In the above define $\Sz(\Delij)$ and $\Szc(\Delij)$ to be the summands corresponding to $(i,j)\in\Zs_{\X}$ and $(i,j)\in\Zcs_{\X}$ respectively.

Now, $\Res(\mb{0})=0$, and so, the idea is to show that there exists $\epsp>0$ such that $\Res(\Del)>0$ for any $\Del$ satisfying $0<\|\Del\|_F<\epsp$. This, with $\eps=\min\{\eps_0,\epsp\}$, will then imply the result (\ref{PLL inequality}). We proceed by dealing with each summand in $\Res(\cdot)$. There are two cases:

\vspace{2mm}

$\mb{[C_1]:}$ Regarding $\Sz(\cdot)$. We have $\Sz(0)=0$, so suppose $\Delij\neq 0$. Then:
\begin{align}
\Sz(\Delij)&>(-\Yij+\Sampij)\Delij + \lam\geq -|\Yij-\Sampij||\Delij|+\lam \nonumber\\[2mm]
&\geq -\cij|\Delij| + \lam, \label{Sz > ...}
\end{align} 
where the last $\geq$ comes from using Lemma \ref{property 2 of F thm} with $\del=\cij>0$. Defining: 
\begin{align}
\eps_1=\frac{\lam}{\max_{(i,j)\in\Zs_{\X}}\ \cij}, \nonumber
\end{align}
which is clearly strictly positive, it follows from (\ref{Sz > ...}) that:
\begin{align}
\Sz(\Delij)>0 \textup{ when } 0<|\Delij|<\eps_1. \nonumber
\end{align}

\vspace{2mm}

$\mb{[C_2]:}$ Regarding $\Szc(\cdot)$. We have $\Szc(0)=0$, so suppose $\Delij\neq 0$. Then, defining: 
\begin{align}
\eps_2=\min_{(i,j)\in\Zcs_{\X}}|\Xij|, \nonumber
\end{align}
which is strictly positive, for any $\Delij$ such that $0<|\Delij|<\eps_1$ it follows that:
\begin{align}
\lam\Ind(\Xij+\Delij\neq 0)-\lam=\lam-\lam=0. \nonumber
\end{align}
Therefore:
\begin{equation}
\Szc(\Delij)=\frac{1}{2}\mu\Delij^2+(-\Yij+\Sampij)\Delij=\frac{1}{2}\mu\Delij^2>0, \nonumber
\end{equation}
where the last equality is due to Lemma \ref{property 1 of F thm}, i.e., $-\Yij=\Sampij$.

\vspace{1mm}

Letting $\epsp=\min\{\eps_1,\eps_2\}$ completes the proof. \qed

\vspace{3mm}

Proving algorithm convergence relies on the following important property of the algorithm map $\Map(\cdot)$:

\newtheorem{lapunov theorem}{Proposition}
\begin{lapunov theorem} \label{lapunov thm}
\textup{Let $\Zoij=\Map(\Xoij)$, and define:
\begin{equation}
\Df(\Xoij,\Zoij)=\PLL(\Zij(\Xoij))-\PLL(\Zij(\Zoij)) \label{lapunov function}
\end{equation}
Then, $\Zoij=\Xoij$ if and only if $\Df(\Xoij,\Zoij)=0$.
}
\end{lapunov theorem}

\textit{Proof}. Clearly, $\Zoij=\Xoij$ implies $\Df=0$. Now, suppose $\Df=0$, in which case:
\begin{align}
\Df(\Xoij,\Zoij,)&=\pllij(\Xoij)-\pllij(\Zoij)=0 \nonumber\\[2mm]
&\Leftrightarrow \pllij(\Xoij)=\pllij(\Zoij) \nonumber\\[2mm]
&\Rightarrow \pllij(\Xoij)=\min_{z}\pllij(z). \label{pll equality}
\end{align}
Letting:
\begin{align}
q(\Xoij)=-\shuroij(\Xoij)^2-2\Yoij\Xoij+1, \nonumber
\end{align}
there are two cases:

\vspace{2mm}

$\mb{[C_1]:}$ $q(\Xoij)\leq 0$. $\pllij(\cdot)$ has a unique minimizer given by $\Zoij$, see (\ref{mij expression 1}) in Theorem \ref{mij is not zero thm}. Thus, (\ref{pll equality}) implies $\Xoij=\Zoij$.

\vspace{2mm}

\begin{figure}[!ht]
\centering
\begin{minipage}{1.0\linewidth}
  \centering
  \includegraphics[width=1\linewidth]{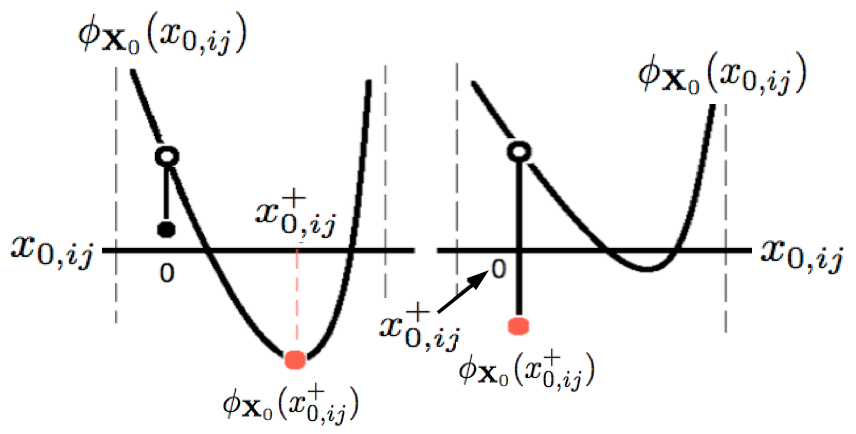}
\end{minipage}
\centering
\begin{minipage}{1.0\linewidth}
  \centering
  \includegraphics[width=1\linewidth]{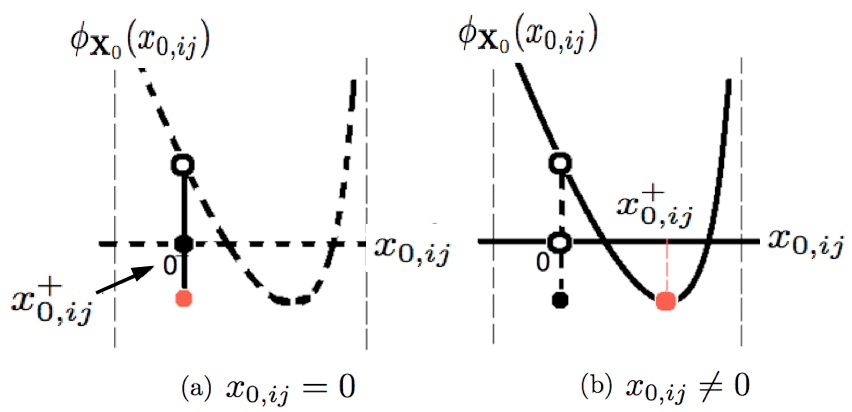}
\end{minipage}
\caption{$\Xoij$ vs. $\pllij(\Xoij)$, where the domain of $\pllij(\cdot)$ contains $0$. (Top) The minimizer is unique, and so, the only time $\pllij(\Xoij)=\min_{z}\pllij(z)$ is when $\Xoij=\Zoij$. (Bottom) (a) The domain of $\pllij(\cdot)$ is only $\{0\}$. This means that $\min_z\pllij(z)$ can only be at $0$. Thus, $\pllij(\Xoij)=\pllij(\Zoij)$ and $\Xoij=\Zoij=0$ at the same time. (b) The domain of $\pllij(\cdot)$ is the entire $\Xoij$-axis except $0$. So, having $\pllij(\Xoij)=\pllij(\Zoij)$ implies $\Xoij=\Zoij$.}
\label{Thm4 C2 fig}
\end{figure}

$\mb{[C_2]:}$ $q(\Xoij)>0$. $\pllij(\cdot)$ has a minimizer given by $0$ and/or by $\Xsij$, where the latter is the unique minimizer of $\pllcij(\cdot)$. Note that $\Xsij\neq 0$ by Theorem \ref{mij is not zero thm}. There are now two subcases:

\hspace{5mm} (i) $\pllij(0)\neq\pllij(\Xsij)$: The minimizer of $\pllij(\cdot)$ is unique, and is either $0$ or $\Xsij$, see expression (\ref{mij expression 2}) and Figure \ref{Thm4 C2 fig} (Top). Therefore, (\ref{pll equality}) implies $\Xoij=\Zoij$.

\hspace{5mm} (ii) $\pllij(0)=\pllij(\Xsij)$: $\pllij(\cdot)$ has two minimizers, $0$ and $\Xsij$, see expression (\ref{mij expression 2}). By the definition of $\Map(\cdot)$, we have: 
\begin{equation}
\Zoij=\Xsij\cdot\Ind(\Xoij\neq 0). \label{boundary value}
\end{equation}
Using (\ref{boundary value}), $\Zoij=0$ implies $\Xoij=0$, and thus, $\Xoij=\Zoij$. If $\Zoij\neq 0$, then $\Xoij\neq 0$ as well. This indicates that $\pllij(\cdot)$ can only have $\Xsij$ as its minimizer, see Figure \ref{Thm4 C2 fig} (Bottom). Thus, (\ref{pll equality}) implies $\Xoij=\Zoij$. \qed

\vspace{3mm}

We have the following sub-sequential result: 

\newtheorem{Map is closed theorem}[lapunov theorem]{Proposition}
\begin{Map is closed theorem} \label{Map is closed thm}
\textup{Assume that (A2) is satisfied. Suppose:
\begin{align}
\left(\Xknij,\Map(\Xknij)\right)\to\left(\Xdij,\Xddij\right) \textup{ as }n\to\infty, \nonumber
\end{align}
Then, $\Xddij=\Map(\Xdij)$. 
}
\end{Map is closed theorem}


\textit{Proof.} Recalling from Theorem \ref{Map is fixed point thm}, when $i=j$, the result easily follows by the continuity of $\Map(\cdot)$. Using the same notation as in Theorem \ref{Map is fixed point thm}, when $i\neq j$, the result also follows if $q(\Xdij)\leq 0$, or $q(\Xdij)>0$ with $\Phi_{\Xd,ij}(\Xdij)\neq 0$. Next, assume $q(\Xdij)>0$ and $\Phi_{\Xd,ij}(\Xdij)=0$. 

Note that $\Xknij$ is an iterate that is thresholded, i.e., it can only be $0$ or $\Xsknbmij\neq 0$ for every $n$, where the latter can only converge to a nonzero number, say, $\Xsddij$. Also note that $\Xknij\to\Xdij$ implies:
\begin{align}
|\Xknspij-\Xknij|\to 0. \label{diff in Xknij}
\end{align}

(i) Suppose $\Xdij=0$. Then for a large enough $N>0$ we must have $\Xknij=0$ for all $n>N$, otherwise (\ref{diff in Xknij}) would be violated. Then, 
\begin{align}
\Map(\Xknij)\to 0 \nonumber
\end{align}
by (A2), and again by the definition of $\Map(\cdot)$, at $\Xdij$ we have:
\begin{align}
\Map(\Xdij)=\Xsij(\Xdij)\cdot\Ind(\Xdij\neq 0)=\Xsij(\Xdij)\cdot 0=\Xddij. \nonumber
\end{align}

\vspace{1mm}
 
(ii) Suppose $\Xdij\neq 0$. Then for a large enough $N>0$ we must have $\Xknij=\Xsknbmij$ for all $n>N$, otherwise (\ref{diff in Xknij}) would be violated. Then, (A2) implies $\Map(\Xknij)=\Xsknij$, which in turn implies:
\begin{align}
\Map(\Xknij)\to\Xsdij, \nonumber
\end{align}
where $\Xsdij=\Xsij(\Xdij)\neq 0$. So, by the definition of $\Map(\cdot)$, at $\Xdij$ we have:
\begin{align}
\Map(\Xdij)=\Xsdij\cdot\Ind(\Xdij\neq 0)=\Xsdij\cdot 1 =\Xddij, \nonumber
\end{align}
completing the proof. \qed

\vspace{3mm}

The remaining results follow from Proposition \ref{Map is closed thm} and require (A1) and (A2). Before proceeding to Proposition \ref{diff iter thm}, two lemmas are needed:

\vspace{1mm}

\newtheorem{lapunov 2 theorem}[quadratic=0 theorem]{Lemma}
\begin{lapunov 2 theorem} \label{lapunov 2 thm}
\textup{Supposing the statement in Proposition \ref{Map is closed thm},
\begin{equation}
\Df(\Xknij,\Map(\Xknij))\to\Dfs \textup{ implies }\Dfs\geq\Df(\Xdij,\Xddij). \label{limit of Df}
\end{equation}
}
\end{lapunov 2 theorem}

\textit{Proof.} Firstly, due to the update of two equal matrix entries at a time, it is obvious that $\Df(\cdot,\cdot)$ is given by:
\begin{equation}
\Df(\Xknij,\Map(\Xknij))=\pllknij(\Xknij)-\pllknij(\Map(\Xknij)). \label{Df simplified 1}
\end{equation}
\indent When $i=j$, $\pllknij(\cdot)$ is continuous w.r.t. its argument, and so, by (\ref{Df simplified 1}) we have $\Dfs=\Df(\Xdij,\Xddij)$.

When $i\neq j$, define:
\begin{equation}
\Indkn=\Ind(\Xknij\neq 0) \textup{ and } \Indmapkn=\Ind(\Map(\Xknij)\neq 0), \nonumber
\end{equation}
and:
\begin{equation}
\Indd=\Ind(\Xdij\neq 0) \textup{ and } \Inddd=\Ind(\Xddij\neq 0). \nonumber
\end{equation}
With these definitions:
\begin{align}
&\Df(\Xknij,\Map(\Xknij)) \nonumber\\[2mm]
&=\left(\pllcknij(\Xknij)-\pllcknij(\Map(\Xknij))\right)+ 2\lam\left(\Indkn-\Indmapkn\right) \nonumber\\[3mm]
&\to \Df(\Xdij,\Xddij)-2\lam\left(\Indd-\Inddd\right)+2\lam\Inds, \label{Df simplified 2}
\end{align}
as $n\to\infty$, where:
\begin{align}
\Inds=\lim_{n\to\infty}\ \Indkn-\Indmapkn \nonumber
\end{align} 
The first two terms in (\ref{Df simplified 2}) result from the continuity of $\pllcknij(\cdot)$ w.r.t. its argument. As a result, in order to show (\ref{limit of Df}), by observation of (\ref{Df simplified 2}), all we have to show is that:
\begin{equation}
\Inds\geq\Indd-\Inddd \label{diff in Ind}
\end{equation}
There are four cases:

\vspace{1mm}

$\mb{[C_1]:}$ Suppose $\Xdij=0$ and $\Xddij=0$. If $\Xknij=0$, then $\Xknij=\Xdij$, and so, $\Map(\Xknij)=\Map(\Xdij)$. But, by Proposition \ref{Map is closed thm} we also have $\Map(\Xdij)=\Xddij$, and thus, 
\begin{align}
\Inds=\Indd-\Inddd. \nonumber 
\end{align}

If $\Xknij\neq 0$, by the definition of the $l_0$ function we have $\Indkn\to 1$. As a result, 
\begin{align}
\Inds\geq 0=\Indd-\Inddd. \nonumber
\end{align}

\vspace{2mm}

$\mb{[C_2]:}$ Suppose $\Xdij\neq 0$ and $\Xddij\neq 0$. Then, $\Indkn\to 1$, and so, 
\begin{align}
\Inds\geq 0=\Indd-\Inddd. \nonumber
\end{align}

\vspace{1mm}

\noindent For the remaining cases, we consider the continuous functions $q(\Xknij)$ and $\Phi_{\Xkn,ij}(\Xknij)$ from (\ref{q function}) and (\ref{Phi function}), respectively.

\vspace{3mm}

$\mb{[C_3]:}$ Suppose $\Xdij=0$ and $\Xddij\neq 0$. If $\Xknij=0$ then $\Xknij=\Xdij$, and so, 
\begin{align}
\Map(\Xknij)=\Map(\Xdij)=\Xddij, \nonumber
\end{align}
where the last equality is due to Proposition \ref{Map is closed thm}. Then we have:
\begin{align}
\Inds=0-1=\Indd-\Inddd. \nonumber
\end{align}

Next, supposing $\Xknij\neq 0$ implies $\Indkn\to 1$. Since $q(\Xdij)=q(0)=1>0$, $\Map(\Xknij)$ is given by (\ref{algorithm map 1}) and by Proposition \ref{Map is closed thm} we also have $\Xddij=\Map(\Xdij)$. So, by the fact that $\Xddij\neq 0$ and the definition of $\Map(\cdot)$, either:
\begin{equation}
\textup{(i) } \Phi_{\Xd,ij}(\Xdij)> 0, \textup{ or } \textup{(ii) } \Phi_{\Xd,ij}(\Xdij)=0 \textup{ and } \Xdij\neq 0 \nonumber
\end{equation}
holds. Clearly, only (i) can be valid in this case, and so, for a large enough $N>0$ we must also have $\Phi_{\Xkn,ij}(\Xknij)>0$ for all $n>N$. Therefore, 
\begin{align}
\Map(\Xknij)=\Xsknij\neq 0 \textup{ for all } n>N. \nonumber
\end{align}
This implies $\Indmapkn\to 1$, and so, 
\begin{align}
\Inds=1-1=0 > -1 = \Indd-\Inddd.\nonumber
\end{align}

\vspace{2mm}

$\mb{[C_4]}$ $\Xdij\neq 0$ and $\Xddij=0$. We firstly have that $\Indkn\to 1$. We cannot have $\Xddij=\Xsdij$, where $\Xsknij\to\Xsdij$, because $\Xsdij\neq 0$. Then, by Proposition \ref{Map is closed thm}, $\Map(\cdot)$ can only be given by (\ref{algorithm map 1}), where from the two resulting possibilities:
\begin{equation}
\textup{(i) } \Phi_{\Xd,ij}(\Xdij)<0, \textup{ or } \textup{(ii) } \Phi_{\Xd,ij}(\Xdij)=0 \textup{ and } \Xdij= 0 \nonumber
\end{equation}
only (i) can be valid. So, for a large enough $N>0$ we must have $\Phi_{\Xkn,ij}(\Xknij)<0$ for all $n>N$, which implies $\Map(\Xknij)=0$ for all $n>N$. Thus, $\Map(\Xknij)\to 0$, which in turn implies $\Indmapkn\to 0$. So, 
\begin{align}
\Inds=1-0=\Indd-\Inddd. \nonumber 
\end{align} \qed

\vspace{3mm}

\newtheorem{bounded sequence theorem}[quadratic=0 theorem]{Lemma}
\begin{bounded sequence theorem} \label{bounded sequence thm}
\textup{The sequence $\{\PLL(\Xk)\}_k$ is bounded from below.
}
\end{bounded sequence theorem}

\textit{Proof.} Firstly, $\|\Xk\|_0>0$. Also, having $\Xk\succ 0$ and $\Samp\succeq 0$ implies $\tr(\Samp\Xk)\geq 0$. As a result, 
\begin{align}
\PLL(\Xk)>-\log\det(\Xk), \nonumber
\end{align}
and by (A1), 
\begin{align}
-\log\det(\Xk)\geq-p\log\alpha, \nonumber
\end{align}
Thus, we have that:
\begin{align}
\PLL(\Xk)>-p\log\alpha \nonumber
\end{align}
which completes the proof.  \qed

\vspace{3mm}

\newtheorem{diff iter theorem}[lapunov theorem]{Proposition}
\begin{diff iter theorem} \label{diff iter thm}
\textup{$\Xkij-\Xkpij\to 0$ as $k\to\infty$.
}
\end{diff iter theorem}

\textit{Proof.} We show the result by establishing a contradiction. So, suppose $\Xkij-\Xkpij\not\to 0$, which means there exists a subsequence:
\begin{equation}
\{\Xij^{k_1}-\Xij^{k_1+1},\Xij^{k_2}-\Xij^{k_2+1},\dots\}\to\del\neq 0. \label{subsequence to del}
\end{equation} 
We note that any subsequence of the sequence in (\ref{subsequence to del}) must converge to $\del$ in order for (\ref{subsequence to del}) to hold. Since the sequence $\{\Xij^{k_1},\Xij^{k_2},\dots\}$ is bounded by (A1), it has at least one limit point. Denote one of these limit points by $\Xdij$ and suppose:
\begin{equation}
\{\Xij^{l_1},\Xij^{l_2},\dots\}\to\Xdij, \label{subsequence l1,l2}
\end{equation}
where 
\begin{align}
\{l_1,l_2,\dots\}\subseteq\{k_1,k_2,\dots\}. \nonumber
\end{align}
Now, consider the sequence $\{\Xij^{l_1+1},\Xij^{l_2+1},\dots\}$, which must have at least one limit point since it is also bounded by (A1). Denote one of these limit points by $\Xddij$, and suppose:
\begin{equation}
\{\Xij^{r_1+1},\Xij^{r_2+1},\dots\}\to\Xddij, \label{subsequence r1+1,r2+1}
\end{equation}
where 
\begin{align}
\{r_1,r_2,\dots\}\subseteq\{l_1,l_2,\dots\}. \nonumber
\end{align}
But now:
\begin{equation}
\{\Xij^{r_1},\Xij^{r_2},\dots\}\to\Xdij, \label{subsequence r1,r2}
\end{equation}
since this sequence is a subsequence of the sequence in (\ref{subsequence l1,l2}). As a result:
\begin{equation}
\{\Xij^{r_1}-\Xij^{r_1+1},\Xij^{r_2}-\Xij^{r_2+1},\dots\}\to\Xdij-\Xddij.  \label{Xdij-Xddij}
\end{equation}
Next, let $\PLLk=\PLL(\Xk)$, and we obviously have $\PLLk\geq\PLLkp$. So, the sequence $\{\PLLk\}_k$ is non-increasing and by Lemma \ref{bounded sequence thm} it must have a finite limit, say, $\PLL^\bullet$. Since:
\begin{align}
\PLLk-\PLLkp\to\PLL^\bullet-\PLL^\bullet=0, \nonumber
\end{align}
by the definition of $\Df(\cdot,\cdot)$ in (\ref{lapunov function}), this means:
\begin{equation}
\Df(\Xkij,\Xkpij)\to 0, \label{Df to 0}
\end{equation}
and so, $\Df(\Xij^{r_n},\Xij^{r_n+1})\to 0$. Then, using Lemma \ref{lapunov 2 thm} we have:
\begin{equation}
0\geq\Df(\Xdij,\Xddij). \label{Df leq 0}
\end{equation}
Since we also have $\Xij^{r_n+1}=\Map(\Xij^{r_n})$, we can use (\ref{subsequence r1+1,r2+1}), (\ref{subsequence r1,r2}) and Proposition \ref{Map is closed thm} to obtain that: $\Xddij=\Map(\Xdij)$. Thus: 
\begin{equation}
\Df(\Xdij,\Xddij)=\Df(\Xdij,\Map(\Xdij))\geq 0. \label{Df geq 0}
\end{equation}
The $\geq$ in (\ref{Df geq 0}) comes from the definition of $\Df(\cdot,\cdot)$ and the fact that:
\begin{equation}
\PLL(\Zij(\Xdij))\geq\PLL(\Zij(\Map(\Xdij))). \nonumber
\end{equation}
As a result, (\ref{Df leq 0}) and (\ref{Df geq 0}) imply $\Df(\Xdij,\Map(\Xdij))=0$, which by Proposition \ref{lapunov thm} means $\Xdij=\Xddij$. Consequently, the limit in (\ref{Xdij-Xddij}) is $0$. Because that sequence is a subsequence of the sequence in (\ref{subsequence to del}) we obtain a contradiction, implying (\ref{subsequence to del}) cannot hold, which completes the proof. \qed

\vspace{3mm}

\newtheorem{limit points are in F theorem}[lapunov theorem]{Proposition}
\begin{limit points are in F theorem} \label{limit points are in F thm}
\textup{$\{\Xkij\}_k$ has limit points, which are all fixed points.
}
\end{limit points are in F theorem}

\textit{Proof.} By (A1), the sequence $\{(\Xkij,\Xkpij)\}_k$ is bounded, and so, has at least one limit point. Denote one of the limit points by $(\Xdij,\Xddij)$. Then, we can find a subsequence $\{(\Xknij,\Xknbpij)\}_n$ such that:
\begin{align}
(\Xknij,\Xknbpij)\to(\Xdij,\Xddij)\textup{ as }n\to\infty. \nonumber
\end{align}
By Proposition \ref{diff iter thm} we have:
\begin{align}
\Xknij-\Xknbpij\to 0, \nonumber
\end{align}
and so, $\Xdij=\Xddij$. Lastly, by Proposition \ref{Map is closed thm} we have $\Xddij=\Map(\Xdij)$, which implies that $\Xdij=\Map(\Xdij)$. \qed

\vspace{2mm}

\newtheorem{general convergence theorem}[lapunov theorem]{Proposition}
\begin{general convergence theorem} \label{general convergence thm}
\textup{$\Xk\to\Fix^\prime$ as $k\to\infty$, where $\Fix^\prime\subseteq\Fix$ is a closed and connected set.}
\end{general convergence theorem}

\textit{Proof.} By Proposition \ref{limit points are in F thm}, $\Fix^\prime\subseteq\Fix$ is the set of limit points of $\{\Xk\}_k$. Then, from Proposition \ref{diff iter thm} we have:
\begin{align}
\Xk-\Xkp\to 0 \nonumber
\end{align}
and $\{\Xk\}_k$ is bounded by (A1). Due to these two facts, we can apply Ostrowski's Theorem 26.1 in \cite[p.173]{Ostrowski73}, which states that the set of limit points of $\{\Xk\}_k$ is closed and connected. \qed 

\vspace{3mm}

\noindent\tb{Proof of Theorem \ref{final convergence thm}:} Define the set of strict local minimizers of $\PLL(\cdot)$:
\begin{align}
\Mset&=\{\Xd: \textup{ there exists }\eps>0 \textup{ such that } \nonumber\\
&\hspace{6mm} \PLL(\Xd)<\PLL(\Xd+\Del), \textup{ for all }0<\|\Del\|_F<\eps\}. \nonumber
\end{align}
This set is derived by considering Theorem \ref{F is local min thm}, by which for $\X\in\Fix$ we have $\X\in\Mset$. This implies $\Mset\neq\emptyset$ and $\Fix\subseteq\Mset$. Since $\Mset$ is the set of distinct local minimizers it must be discrete i.e. consists only of isolated points. If not, there exists a connected subset which is a continuum and this violates the strict inequality. Therefore, the subset $\Fix$ is a discrete set as well. However, by Proposition \ref{general convergence thm} the limit point set of $\{\Xk\}_k$ is a connected subset of $\Fix$. Hence, the limit point set must contain only a single point, say $\Xd$, and the result follows. \qed

\bibliographystyle{IEEEtran}
\bibliography{Ref}

\end{document}